\documentclass{bmvc2k}
\usepackage{array}
\usepackage{amssymb}
\usepackage{placeins}
\usepackage{xcolor,soul}
\usepackage[framemethod=tikz]{mdframed}
\usepackage[ruled,vlined]{algorithm2e}
\usepackage{multirow} 
\usepackage{caption} 
\usepackage{subcaption}
\usepackage{ wasysym }
\usepackage{graphicx}
\usepackage{booktabs} 
\usepackage[pagebackref]{hyperref} 
\usepackage[activate={true,nocompatibility},final,tracking=true,kerning=true,spacing=true,factor=1100,stretch=10,shrink=9]{microtype}
\usepackage[accsupp]{axessibility}  

\newcolumntype{C}{>{\centering\arraybackslash}p{0.06\textwidth}}

\title{SVS: Adversarial refinement\\ for sparse novel view synthesis}

\addauthor{Violeta Menéndez González}{v.menendezgonzalez@surrey.ac.uk}{1,2}
\addauthor{Andrew Gilbert}{a.gilbert@surrey.ac.uk}{1}
\addauthor{Graeme Phillipson}{graeme.phillipson@bbc.co.uk}{2}
\addauthor{Stephen Jolly}{stephen.jolly@bbc.co.uk}{2}
\addauthor{Simon Hadfield}{s.hadfield@surrey.ac.uk}{1}
\addinstitution{
 Centre for Vision, Speech and Signal Processing (CVSSP)\\
 University of Surrey\\
 Guildford, UK
}
\addinstitution{
 BBC R\&D\\
 MediaCityUK\\
 Salford, UK
}

\runninghead{Menendez Gonzalez \etal}{Sparse Novel View Synthesis}

\def\etal{\emph{et al}\bmvaOneDot}

\begin{document}

\maketitle

\begin{abstract}
This paper proposes Sparse View Synthesis. This is a view synthesis problem where the number of reference views is limited, and the baseline between target and reference view is significant.
Under these conditions, current radiance field methods fail catastrophically due to inescapable artifacts such 3D floating blobs, blurring and structural duplication, whenever the number of reference views is limited, or the target view diverges significantly from the reference views. 

Advances in network architecture and loss regularisation are unable to satisfactorily remove these artifacts. The occlusions within the scene ensure that the true contents of these regions is simply not available to the model.
In this work, we instead focus on hallucinating plausible scene contents within such regions. To this end we unify radiance field models with adversarial learning and perceptual losses. The resulting system provides up to 60\% improvement in perceptual accuracy compared to current state-of-the-art radiance field models on this problem.
\end{abstract}

\begin{figure*}[htbp]
    \begin{center}
        \begin{subfigure}[b]{0.25\textwidth}
             \includegraphics[width=\textwidth]{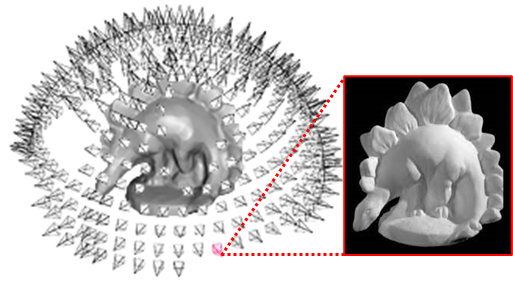}
             \caption{Dense View Synthesis}
             \label{fig:teaser1}
        \end{subfigure}
        \hspace{1cm}
        \begin{subfigure}[b]{0.25\textwidth}
             \includegraphics[width=\textwidth]{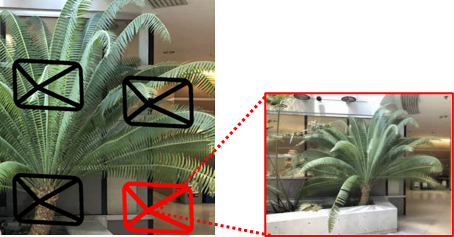}
             \caption{Few View Synthesis}
             \label{fig:teaser2}
        \end{subfigure}
        \hspace{1cm}
        \begin{subfigure}[b]{0.25\textwidth}
             \includegraphics[width=\textwidth]{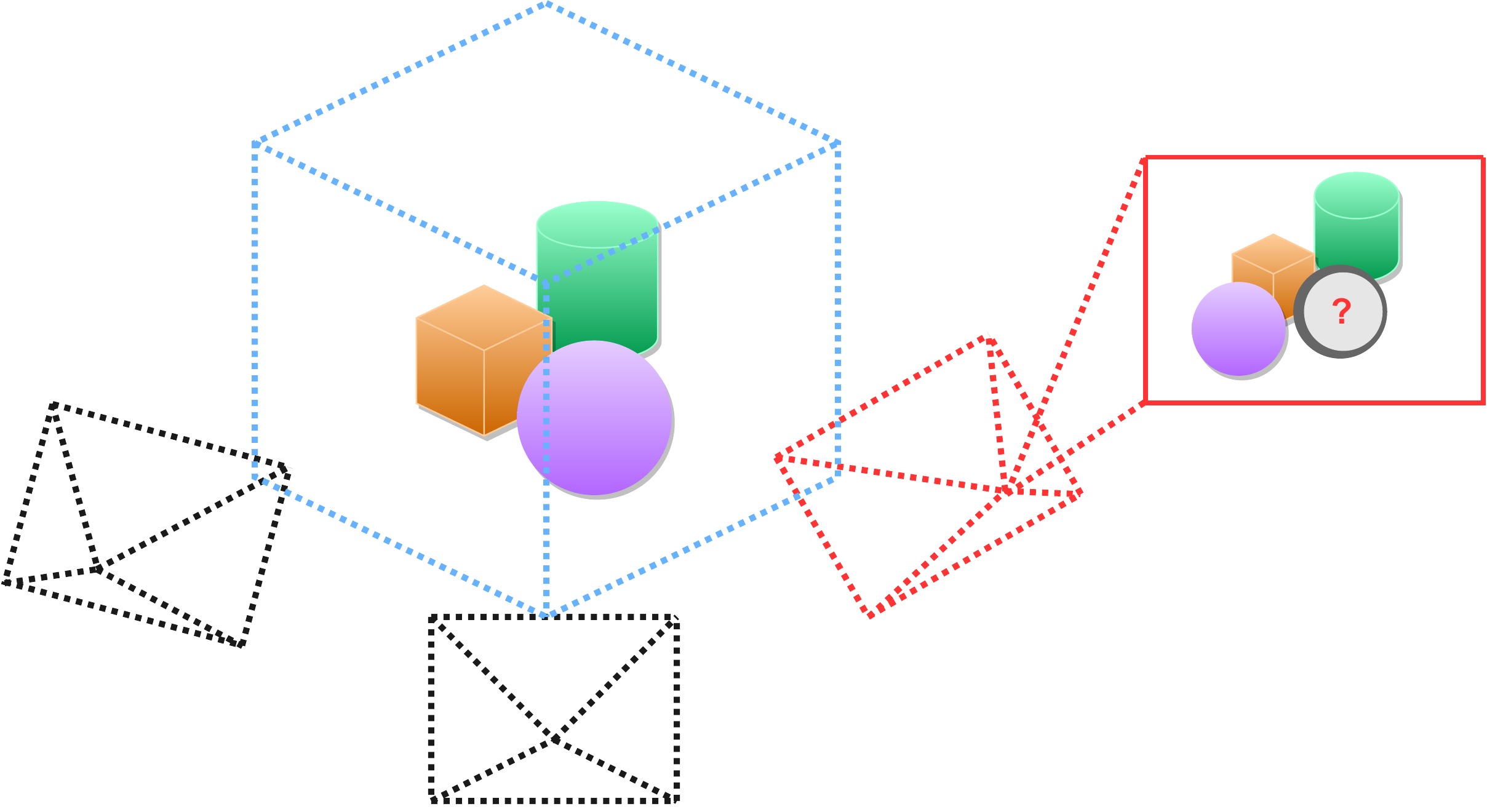}
             \caption{Sparse View Synthesis}
             \label{fig:teaser3}
        \end{subfigure}
    \end{center}
    \caption{Different view synthesis operating modes, with varying numbers of views and varying baselines between views. In each case the black cameras are reference views and the red camera is the target view. Note that with few views and a wide baseline, occluded regions appear in the rendered scene which are visible in only 1 or none of the reference views.}
    \label{fig:teaser}
\end{figure*}

\section{Introduction}
\label{sec:intro}

Novel view synthesis (NVS) is the problem of generating new camera viewpoints of a scene, given a fixed set of views of the same scene.
Most modern NVS methods approach the problem as that of learning a generative model for the scene, conditioned on the camera pose.
Key challenges with current NVS approaches are inferring the scene's 3D structure given a restricted set of reference views, which are not necessarily coplanar with the target view. We call this the Sparse View Synthesis problem, and it raises significant challenges with the inpainting of occluded and unseen parts of the scene. This task has wide applications in image and video editing, Virtual Reality, or as a pre-processing step for other computer vision and robotics tasks. This makes novel view synthesis a key problem in modern computer vision.

Recent years have seen rapid growth in this field. Most notably, neural rendering approaches like Neural Radiance Fields (NeRF) and its advancement \cite{mildenhall2020nerf,martin-brualla2021} have become very popular due to their photo-realistic results. However, these approaches tend to be very expensive, requiring a multitude of input views and a very long per-scene optimization process to obtain high-quality radiance fields. While this can be useful for tasks such a 3D object reconstruction for graphics design, it is far from practical and accessible for other applications such as live event capture.

This work aims to make neural scene reconstruction more accessible and applicable to real world scene capture. In particular we propose a method which does not require scene-specific model training, while still providing realistic results from a small sparse set of input views. We refer to this problem as Sparse View Synthesis. The key challenge is effectively recognizing and handling occluded areas, which were not observed from the small number of training views, while keeping rendering efficient. This necessitates a greater focus on generalization and extrapolation and pure synthesis, as opposed to the data aggregation of traditional radiance field models.

Some methods have approached this generalisation problem by reconstructing geometry priors. Indeed models like \cite{chibane2021srf,chen2021mvsnerf} attempt to replicate classic multi-view stereo behaviour using deep learning techniques. However, these approaches have focused on narrow baseline extrapolation, where occlusions are limited.

To be able to deal with occlusions and artefacts sensibly, we unify adversarial training with radiance field models (fig.~\ref{fig:teaser}). The adversarial training paradigm was first introduced as Generative Adversarial Networks~\cite{goodfellow2014gans}. This was designed to help enrich the output variability of generative models, while dealing with artefacts in a realistic way. In the domain of neural radiance fields, this has the potential to ensure realistic extrapolation in unobserved regions. We have made our code publicly available\footnote{\url{https://github.com/violetamenendez/svs-sparse-novel-view}}.

\section{Background}
Classical approaches to novel view synthesis (also known as Image-based rendering (IBR) \cite{mcmillan1995,debevec1998,chaurasia2013}) have typically relied on restrictive intermediate representations of geometry. These range from multi-layer representations like Plane Sweep Volumes~\cite{flynn2016deepstereo,xu2019}, Multi-Plane Images (MPI)~\cite{zhou2018stereomag,flynn2019deepview}, or Layered Depth Images (LDIs)~\cite{shade1998,shih2020} to more complex voxel grids~\cite{sitzmann2019deepvoxels,shi2021} and 3D point clouds~\cite{wiles2020synsin, xu2022}.

More recently, NeRF~\cite{mildenhall2020nerf} proposed an entirely neural scene representation, where a Multi-Layer Perceptron (MLP) parameterises a volumetric function which maps position and viewing direction to density and colour.
Unfortunately, in it's original form NeRF is very costly to run and has to be optimised per scene, which prevents it from being useful in many important applications. Subsequent approaches~\cite{martin-brualla2021,barron2021mipnerf,srinivasan2021} have tried to loosen these constraints or improve performance~\cite{hu2022}. Despite this, all these approaches struggle to generalise across scenes, require dense input images, and are very costly to run. In particular recent works have focused on introducing additional data augmentation~\cite{chen2022a} and regularisation systems~\cite{kim2022, rebain2022lolnerf, niemeyer2022regnerf, deng2022} to reduce the number of viewpoints required to build a scene-specific NeRF model.

To overcome the limitations of the scene-specific implicit representation, some approaches have attempted to combine the geometry learning strengths of IBR approaches with the power of neural rendering techniques. 
IBRNet~\cite{wang2021ibrnet} aggregates 2D feature information from source views along a given ray to compute its final colour. SRF~\cite{chibane2021srf} emulates classical stereo matching techniques by learning an ensemble of pair-wise similarities. 
But the results are very blurry, cannot handle specularities, and the model is very expensive to run. PixelNeRF~\cite{yu2021pixelnerf} manages to generalise to new scenes using as few as one input image and no explicit geometry-aware 3D structures. However, it tends to overfit to the training set, failing to generalise well.On the other hand, MVSNeRF~\cite{chen2021mvsnerf} reconstructs an encoding volume based on a 3D feature Plane Sweep Volume~\cite{flynn2016deepstereo}. This model works on only three input images and is generalisable to different scenes.
Further developments were made based on geometric constraints~\cite{johari2022geonerf} and recurrent aggregation~\cite{zhang2022nerfusion}.
However, in all these systems only the scene content visible from the reference view is well reconstructed. The outputs contain significant artefacts in challenging or occluded regions which require further fine-tuning per scene. 
These techniques also lack any mechanism to generate image content in areas which are occluded in all inputs. This becomes a significant problem in Sparse View Synthesis problems, where the target view is not closely aligned with the reference view.

With the development of \textit{Generative Adversarial Networks} (GANs)~\cite{goodfellow2014gans}, it has become possible to generate novel photo-realistic content \cite{karras2019stylegan,karras2020stylegan2,choi2018stargan,choi2020starganv2}. Several works have applied adversarial methods to the controllable novel view synthesis of objects. \textit{HoloGAN}~\cite{nguyen-phuoc2019hologan} learns object representations extracting 3D features from single natural images and disentangles shape and appearance.
GRAF~\cite{schwarz2020graf} achieves disentanglement of object properties while not requiring 3D supervision. All single-view methods base their 3D representations on a single 2D image, which suffer from single-view spatial ambiguities. Nanbo \etal~\cite{nanbo2020} address this by trying to composite multi-object scenes leveraging multiple views. GIRAFFE~\cite{niemeyer2021giraffe} incorporates compositional 3D scene structure to the model to handle multi-object scenes. Pix2NeRF~\cite{cai2022pix2nerf} trained a generator system to produce random NeRF volumes which could then be combined with a decoder for GAN inversion. GNeRF~\cite{meng2021gnerf} uses adversarial training to reconstruct NeRFs with unknown camera poses. pi-GAN~\cite{chan2021pigan} models partial single objects using periodic activation functions. All of these models aim to disentangle image composition for scene editing, or are limited to simple scenes comprised of one or a few simple objects. DeVris \etal~\cite{devries2021} decompose complex scenes in many local specialised Radiance Fields. This requires additional depth information and extremely expensive training. Our method on the other hand leverages adversarial training to achieve photo-realistic image generation of unconstrained occluded areas in Sparse View Synthesis.

\section{Approach}

\begin{figure*}[htbp]
    \begin{center}
        \includegraphics[width=\linewidth]{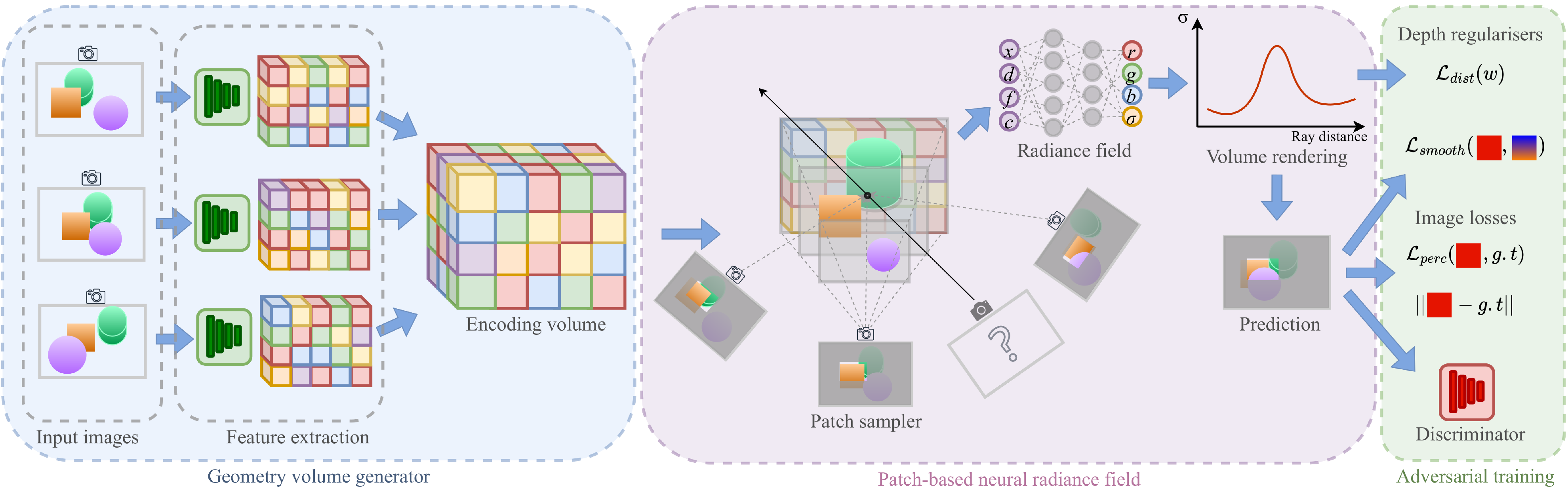}
    \end{center}
    \caption{\textbf{Model overview.} }
    \label{fig:overview}
\end{figure*}

We propose a pipeline based on a Plane Sweep Volume~\cite{flynn2016deepstereo} neural encoding following MVSNeRF~\cite{chen2021mvsnerf}. From this volume we sample random patches using radiance fields~\cite{mildenhall2020nerf} which are supervised by an adversarial loss. As opposed to \cite{mildenhall2020nerf}, we don't require a dense set of input images. We aim to learn a general model that can be applied to new unseen scenes without fine-tuning. Our model also aims to handle significant occlusions due to large baseline changes from sparse input viewpoints. In particular, we train a generalisable adversarial framework for radiance fields. An overall visualisation of our proposed model can be seen in Figure \ref{fig:overview}.

Given some sparse input images our model reconstructs an embedded neural volume which allows the model to reason about the implicit geometry of a scene. We use ray marching to sample from this volume and render a new point of view. We leverage adversarial training to help provide plausible rendering for large dis-occlusions and artefacts that arise from the large baseline changes. The following sections will detail each of these elements of our approach in turn.

\subsection{Geometry volume generator}

As the initial encoder for our generator, we use a 3D CNN encoding volume~\cite{chen2021mvsnerf} which integrates 2D CNN features of the input images. This allows the network to extract correlations between images, which can then be used to reason about geometry. The focus on image correlations as a mechanism for geometry extraction helps the network generalise to previously unseen scenes. The encoding volume is created at the reference view by warping multiple sweeping planes of source view features. This is in contrast to techniques like Deep Stereo~\cite{flynn2016deepstereo}, which perform plane sweeps using the raw colour pixels to produce their correlation volume.

To construct this volume, we first extract the deep features $\{\mathbf{F}_{i} \mid \mathbf{F}_{i} \in \mathbb{R}^{\frac{H}{4} \times \frac{W}{4} \times C}\}_{i=1}^{N}$ of the \emph{N} input images $\{\mathbf{I}_{i}\mid \mathbf{I}_{i} \in \mathbb{R}^{H \times W \times 3}\}_{i=1}^{N}$ using a deep 2D convolutional network $\mathbf{F}_i = E(\mathbf{I}_i | \mathbf{w}_E)$. This network consists of downsampling convolutional layers, batch-normalization and ReLU activation layers. For efficiency and generality, the feature encoding network is shared across all views~\cite{yao2018mvsnet}.

Next we must align each feature map $\{\mathbf{F}_i\}$ to the reference view at multiple depths to encode the plane sweep volume. To achieve this, a homography $\mathcal{H}_i\left(d\right)$ is computed for each view at each depth.
Given the camera parameters $\{\mathbf{K}_i,\mathbf{R}_i,\mathbf{t}_i\}$ (intrinsics, rotation and translation) for camera $i$ the homography is defined as 
\begin{equation}
    \mathcal{H}_i\left(d\right)=\mathbf{K}_i \cdot \mathbf{R}_i \cdot \left(\mathbf{I}+\frac{\left(\mathbf{t}_{ref}-\mathbf{t}_i\right)\cdot\mathbf{n}_{ref}^{T}}{d}\right)\cdot\mathbf{R}_{ref}^{T}\cdot \mathbf{K}_{ref}^{T}
\end{equation}
where $\mathbf{I}$ is the $3\times3$ identity matrix, $\mathbf{n}_{ref}$ the principle axis of the reference camera, and $d$ is the depth which the images are being warped to. This operation is differentiable, which allows for end-to-end training of the feature encoding network weights $\mathbf{w}_E$ based on the downstream reconstruction losses.

Applying this homography to the feature maps gives us the warped feature sweep volumes
\begin{equation}
    \mathbf{V}_i=\{\mathbf{F}_i \cdot \mathcal{H}_i\left(d\right) \mid \forall d=1,...,D\}.
\end{equation}
Then, a cost volume $\mathbf{C}$ is created by aggregating all the warped feature sweep volumes, which encode appearance variations across views. To do this, a variance based cost metric is used, as it allows to use an arbitrary number of input views,
\begin{equation}
    \mathbf{C} = Var\left(\mathbf{V}_i\right) = \frac{\sum_{i=1}^{N}\left(\mathbf{V}_i-\overline{\mathbf{V}}_i\right)^{2}}{N}.
\end{equation}

This cost volume is then processed using a 3D CNN UNet-like network~\cite{ronneberger2015unet}. This includes downsampling and upsampling layers with skip connections, to propagate scene appearance information. The output of this network is the neural embedding volume $\mathbf{E}=V(\mathbf{C}|\mathbf{w}_V)$. This embedding volume represents the feature correlations from the point of view of the reference frame's plane sweep volume. The structure of this volume is consistent across any arrangement of input viewpoints, and even any number of input views. This allows the system to generalize to new scene arrangements.

\subsection{Volume rendering}

\begin{figure*}[htbp]
    \begin{center}
        \includegraphics[width=\linewidth]{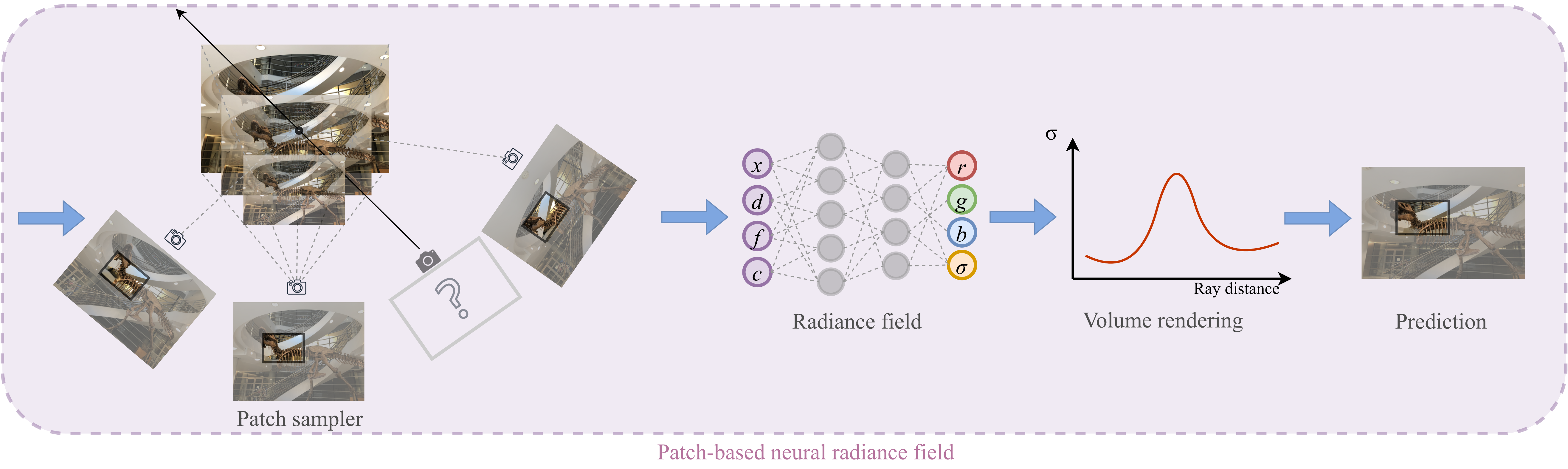}
    \end{center}
    \caption{\textbf{Volume rendering pipeline}}
    \label{fig:volume_rendering}
\end{figure*}

We next use a neural radiance MLP with parameters $\mathbf{w}_\Theta$ to decode the embedding volume into volume density and view-dependent radiance (colour). Given a 3D point $x$, and a viewing direction $d$, we optimise a network $F_{\Theta}$ to regress the density $\sigma$ and colour $r$ from the volume $\mathbf{E}$ at that point $x$. To allow the correlations and structures in $\mathbf{E}$ to be mapped back to the original scene albedo, we use the pixel colour of the original image inputs $\mathbf{I}$ as additional conditioning information.
\begin{equation}
    F_{\Theta}\colon \left(x,d,\mathbf{E},\mathbf{I}|\mathbf{w}_\Theta\right) \mapsto \left(\sigma_{x,d}, r_{x,d}\right)
\end{equation}

We use differentiable ray marching to regress the colour of reference image pixels. This is done by projecting (``marching'') a ray
through a pixel $p$ in the reference image $\mathbf{I}_{ref}$. We can use the neural radiance network to obtain the radiance $r_\gamma$ and density $\sigma_\gamma$ at regular intervals $\gamma \in [1..\inf]$ along this ray via
\begin{equation}
    \left(\sigma_{\gamma}, r_{\gamma}\right) = F_{\Theta}\left(\mathbf{t}_{ref} + \gamma\hat{d},\hat{d},\mathbf{E},\mathbf{I}|\mathbf{w}_\Theta\right)
\end{equation}
where $\hat{d} = \mathbf{R}_{ref}^{T}\cdot \mathbf{K}_{ref}^{T}p$.
We can use these regular samples from $F_{\Theta}$ to obtain the predicted colour of the pixel $R(p)$ via volume rendering equation~\cite{kajiya1984}:
\begin{align}
    \label{eq:ray_render}
    R(p) &= \sum_{\gamma} \tau_\gamma \left(1-exp\left(-\sigma_\gamma\right)\right)r_\gamma \\
    \tau_\gamma &= exp\left(-\sum_{j=1}^{\gamma-1}\sigma_j\right)
\end{align}
where $\tau_\gamma$ is the transmittance at sample $\gamma$, which represents the probability that the ray travels up to $\gamma$ without hitting another particle.

It is intuitive that the proposed approach will be able to predict density based on the consistency of feature representations between views. We can even see how analysing exactly which views correlate well for a given point can provide hints about occlusions, and guidance for albedo lookup. However, there is no simple mechanism to distinguish a region which has low correlation due to being empty, and one with low correlation due to being occluded in all views. The prevalence of these fully occluded regions grows drastically as the number of input views is reduced, and leads traditional radiance field models to produce reconstructions full of unrealistic holes.

\subsection{Adversarial training}
To combat this, we couple the above Generator network with a Discriminator network and undertake adversarial training. This makes it possible to  enforce realism in unobserved regions. However, effective adversarial training requires spatial structure in the generated output, therefore we use a patch based neural generator function based on equation~\ref{eq:ray_render}. The use of a patch based generator serves two purposes. Firstly, it exponentially increases the number of possible training samples, ensuring that the discriminator is not able to memorize the training dataset. Secondly it greatly improves training efficiency as it can be expensive to repeatedly render entire images via the Neural Radiance Field.

Following Schwarz \etal~\cite{schwarz2020graf}, we generate a variable patch that scales with training time. This allows for a variable receptive field. The patch $\mathbf{P}_p$ centred on pixel $p$ of size $\delta\times\delta$ is defined as 
\begin{equation}
    \mathbf{P}\left(p,s\right)=\left\{\left(si+p_x,sj+p_y\right)\mid i,j \in \left\{-\frac{\delta}{2},...,\frac{\delta}{2}\right\}\right\}
\end{equation}
where $s$ is the scale that controls the active field of the patch. The scale exponentially decays during the training process, allowing our convolutional discriminator to learn independently of the image resolution.

\begin{figure}[htbp]
	\centering
	\begin{subfigure}[t]{0.3\textwidth}
  		\centering
  		\includegraphics[width=\textwidth]{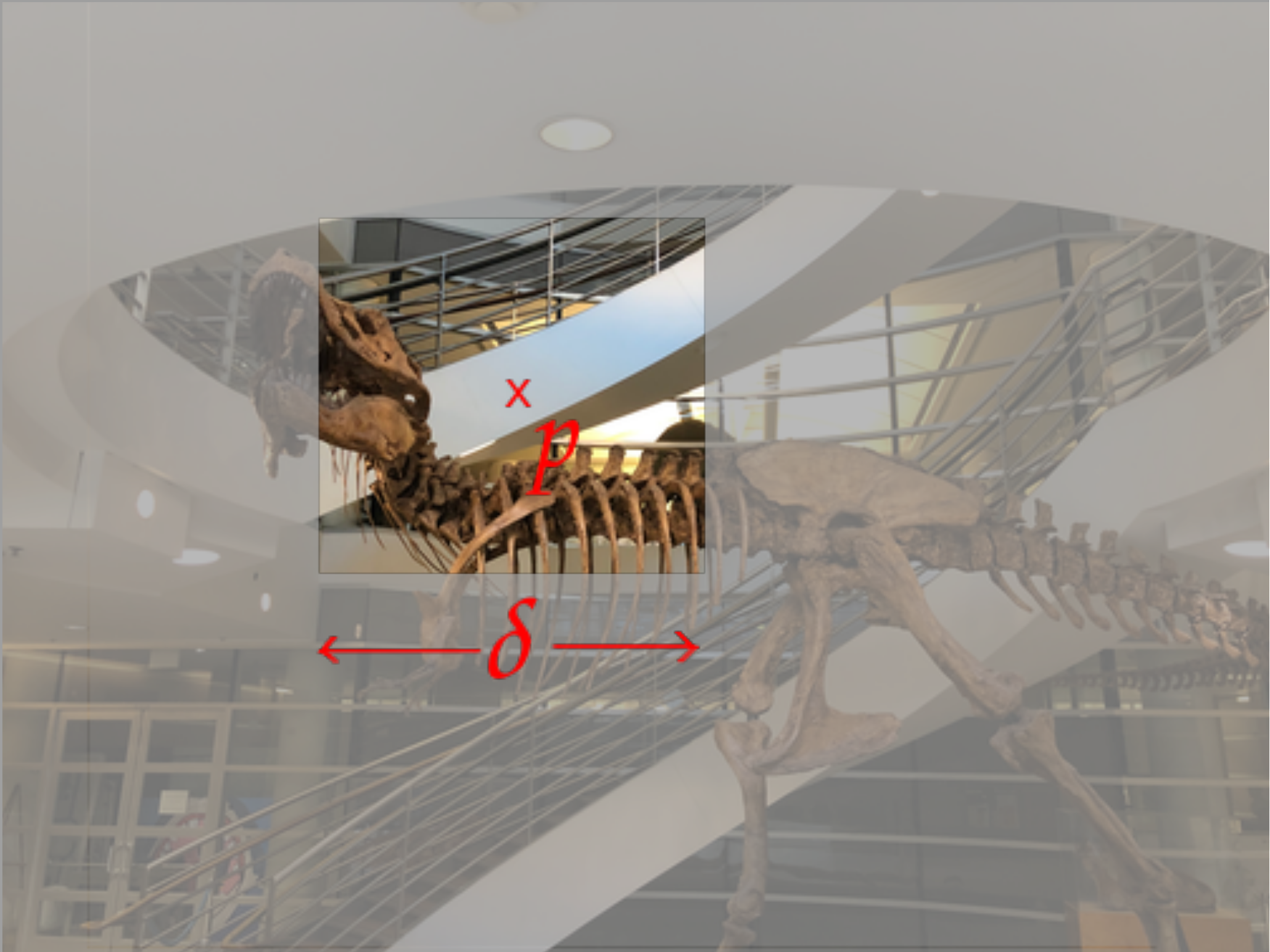}
  		\label{subfig:patch1}
	\end{subfigure}
	~
	\begin{subfigure}[t]{0.3\textwidth}
  		\centering
  		\includegraphics[width=\textwidth]{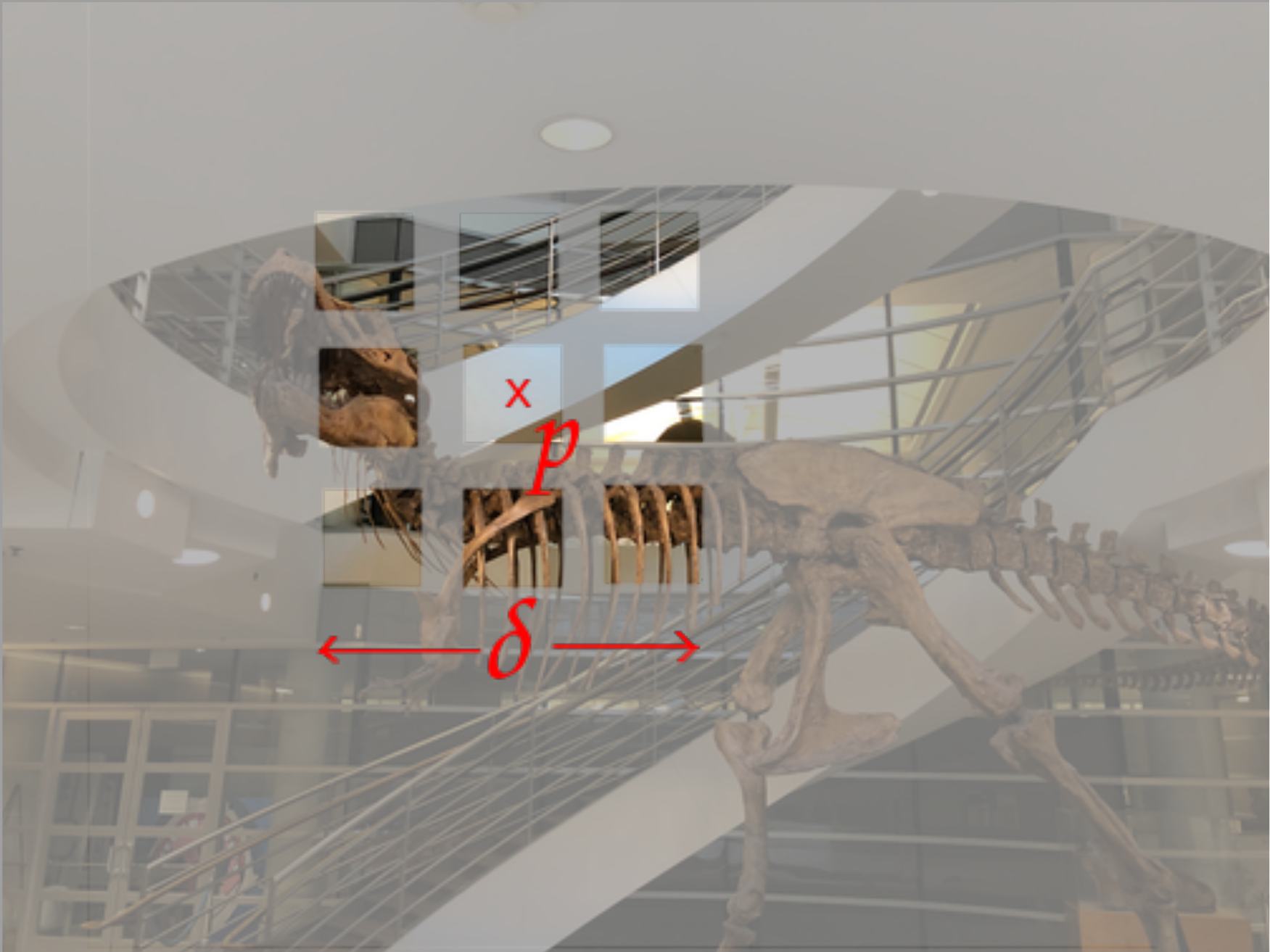}
  		\label{subfig:patch3}
	\end{subfigure}
	~
	\begin{subfigure}[t]{0.3\textwidth}
  		\centering
  		\includegraphics[width=\textwidth]{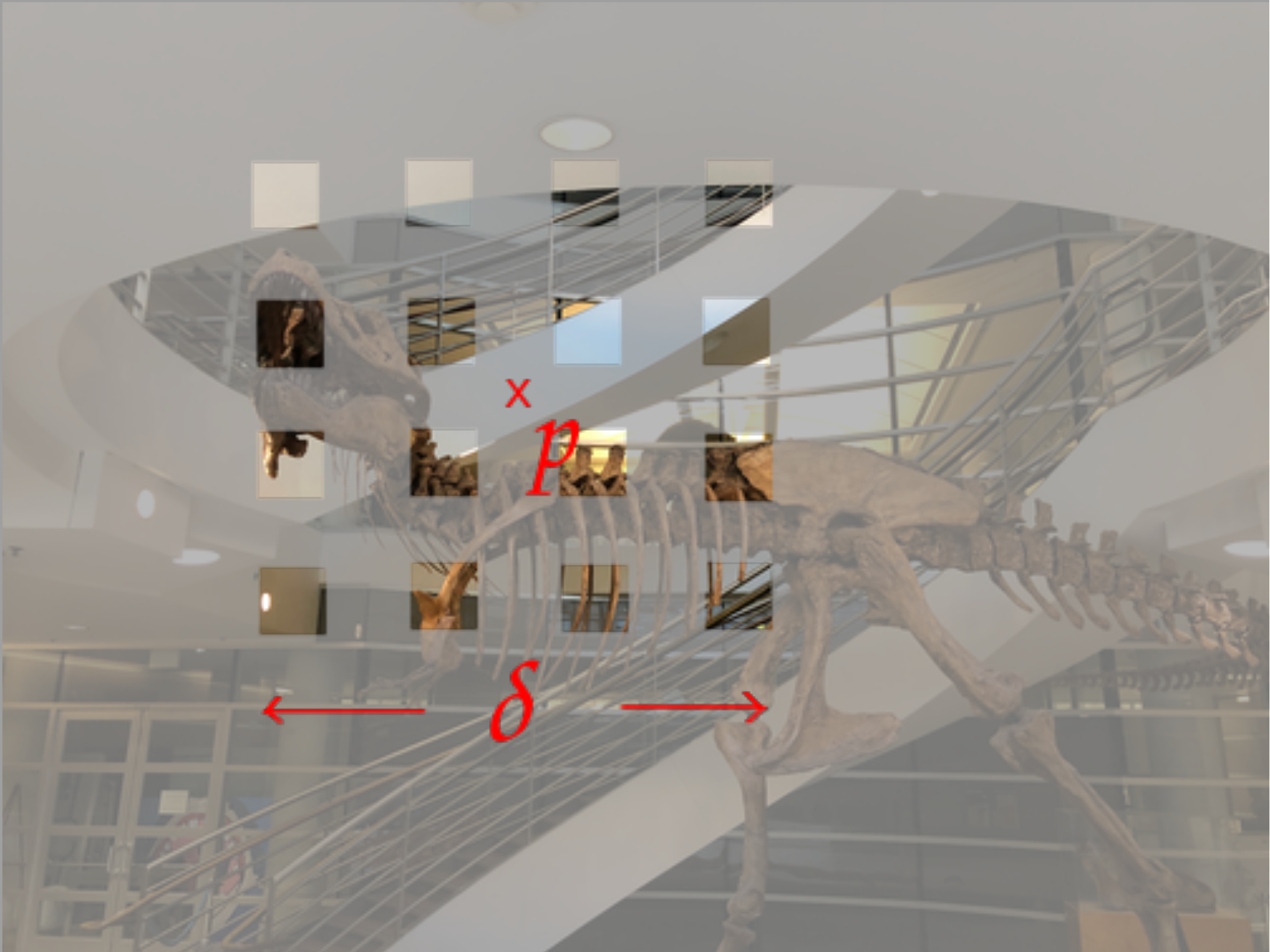}
  		\label{subfig:patch16}
	\end{subfigure}
	\captionsetup{skip=3pt,width=.8\textwidth}
	\caption{\textbf{Patches.} An illustration of the varying patch scale over the course of training.}
	\label{fig:patches}
\end{figure}

Given the ground truth patch $\mathbf{P}_p^*$ for the target view, we compute the $L_1$ loss between a randomly selected real and target patch
\begin{equation}
    \mathcal{L}_{rec} = ||\mathbf{P}_p-\mathbf{P}_p^*||
    \text{  where  }
    p\sim \mathcal{U}([0,0], [W,H]).
\end{equation}
This $L_1$ loss is effective at enforcing low-frequency correctness in the output. However, it can lead to overly-blurred results and difficulty recovering high-frequency structures. We therefore follow an LSGAN~\cite{mao2017lsgan} approach and augment this with a convolutional PatchGAN~\cite{isola2017patchgan} discriminator network $D_\Phi$ with parameters $\mathbf{w}_\Phi$.

The discriminator network takes patches as input, and is trained to classify input patches from the ground truth and from the generator as real or fake respectively. As such the discriminator loss is defined as
\begin{equation}
   \mathcal{L}_D = 
   ||1-D_{\Phi}\left(\mathbf{P}_p^*\right)||_{2}^{2}+
   ||D_{\Phi}\left(\mathbf{P}_p\right)||_{2}^{2}
    \text{  where  }
    p\sim \mathcal{U}([0,0], [W,H]).
\end{equation}

Finally, we can use the PatchGAN discriminator's loss to also create an adversarial loss which constrains the generator network
\begin{equation}
    \mathcal{L}_{G} = 
    \lambda||1-D_{\Phi}\left(\mathbf{P}_p\right)||_{2}^{2}
    \text{  where  }
    p\sim \mathcal{U}([0,0], [W,H]).
\end{equation}
where $\lambda$ is a weighting factor.
This adversarial loss encourages the generator to produce more realistic outputs which are able to fool the discriminator. Importantly, any holes in the reconstruction due to occlusions will provide obvious clues for the discriminator. Therefore the generator can only succeed in its task if the holes are filled with hallucinated photo-realistic content. 

We should re-iterate that this entire pipeline is fully differentiable. This includes the discriminator, the patch based volumetric rendering, the radiance field estimation, the feature correlation computation, the homographic plane-sweep warping and the input image feature lookup. As such, our adversarial loss $\mathcal{L}_{G}$ is able to constrain all the learnable parameters $\mathbf{w}_E, \mathbf{w}_V, \mathbf{w}_\Theta$ apart from those in the discriminator $\mathbf{w}_\Phi$ which are optimised based on $\mathcal{L}_{D}$. These two optimizations are performed using separate Adam optimisers~\cite{kingma2014adam} which are alternated.

\subsection{Depth regularisation}
Because Sparse View Synthesis is an ill-defined problem, we found that the predicted depth images were extremely noisy, if not completely nonsensical. To be able to reconstruct a well defined scene geometry, the predicted depth needs to be coherent with the image. We approach this issue by adding two depth regularisers.

\subsubsection{Edge-aware depth smoothness}
Firstly, we introduce a depth smoothness loss to encourage the network to generate continuous surfaces, similar to monocular depth estimation approaches~\cite{godard2017monodepth}. As depth discontinuities usually happen at colour edges~\cite{heise2013pmhuber}, the depth smoothness is weighted with the colour image gradients $\partial I$.
\begin{equation}
    \label{eq:dept_smooth}
    \mathcal{L}_{smooth}\left(d\right) = \frac{1}{N}\sum_{i,j}\left[ \left|\partial_xd_{ij}\right|\mathrm{e}^{-||\partial_xI_{ij}||}+\left|\partial_yd_{ij}\right|\mathrm{e}^{-||\partial_yI_{ij}||}\right]
\end{equation}
where $d_{i,j}$ is the predicted depth at pixel $(i,j)$, and $I_{i,j}$ the respective colour value.

\subsubsection{Distortion loss}
In addition to smooth surfaces, we also want to get rid of other potential artefacts like ``floaters'' (small disconnected regions of occupied space that look fine from the input views, but wouldn't be coherent if seen from another view), and ``background collapse'' (far surfaces modeled as semi-transparent clouds of dense content in the foreground). NeRF-based models~\cite{mildenhall2020nerf} try to achieve this by adding Gaussian noise to the output $\sigma$ values during optimisation. But this does not eliminate all geometry artefacts, and reduces the reconstruction quality. Instead, we follow Barron~\etal~\cite{barron2022mipnerf360} and include a distortion loss in our regularisation.
\begin{align}
    \mathcal{L}_{dist}\left(\mathbf{w}\right) &= \sum_{i,j}\mathrm{w}_i \mathrm{w}_j + \frac{1}{3}\sum_{i}\mathrm{w}^{2}_i \\
    \mathrm{w}_i &= \tau_{i} \left(1-exp\left(-\sigma_{i}\right)\right)
\end{align}
where $\mathrm{w}_i$ are the alpha compositing weights at ray sample $i$, derived from equation \ref{eq:ray_render}. This regulariser minimises the weighted distances between all pairs of ray points, and the weighted size of each individual point. This helps the distribution function of the density along the rays approximates a delta function.
Finally, we combine all the generator losses and regularisers with an LPIPS~\cite{zhang2018lpips} perceptual loss.
Our total loss is as follows:
\begin{equation}
    \mathcal{L}_{total} = \frac{1}{2}\mathcal{L}_{D} + \mathcal{L}_{G} + \lambda_{rec} \mathcal{L}_{rec} + \mathcal{L}_{perc} + \lambda_{smooth} \mathcal{L}_{smooth} + \lambda_{dist} \mathcal{L}_{dist}
\end{equation}
where we chose $\lambda_{rec}=20,\lambda_{smooth}=0.4,\lambda_{dist}=0.001$ for our experiments.


\section{Experimental setup}

For training we are using two different commonly used datasets, DTU~\cite{jensen2014dtu} and Forward-Facing (LLFF) data~\cite{mildenhall2019llff}. The DTU dataset consists of a variety of scenes and objects taken in a lab setup. We follow the same training approach in related papers~\cite{yu2021pixelnerf,chen2021mvsnerf}, and split the dataset into 88 training scenes and 16 testing scenes, using an image resolution of 512$\times$640. The Forward-Facing dataset consists of handheld phone captures taken in a 2D grid. We split the dataset into 35 training sets and 8 for testing in the same scenes used for NeRF. Because we focus on the sparse view synthesis problem, models are trained on 3 input views per scene.

\subsection{Baseline models}

We compare our method against the current state-of-the-art neural rendering methods for Sparse View Synthesis. All methods are trained over LLFF and DTU using three input images. We evaluate IBRNet~\cite{wang2021ibrnet}, MVSNeRF~\cite{chen2021mvsnerf} and our method over long baseline movements. We weren't able to train GeoNeRF~\cite{johari2022geonerf} as the code hasn't been released yet, and the results in their paper are for a much easier problem.

\subsection{Evaluation of accuracy}

For the purpose of quantifying how well our model performs, we make use of several popular metrics that measure different characteristics of an image. To measure image quality, we use Peak Signal-To-Noise Ratio (PSNR)~\cite{huynh-thu2008psnr} and Structural SIMilarity (SSIM)~\cite{wang2004ssim} index. PSNR shows the overall pixel consistency, while SSIM measures the coherence of local structures. These metrics assume pixel-wise independence, which may assign favourable scores to perceptually inaccurate results. For this reason, we also include the use of a Learned Perceptual Image Patch Similarity (LPIPS)~\cite{zhang2018lpips} metric, which aims to capture human perception using deep features.

\begin{table}[htpb]
    \captionsetup{skip=3pt,width=.9\textwidth}
	\caption{\textbf{Quantitative evaluation.} We evaluate our model over the DTU and Forward Facing datasets.  \textbf{Bold} is best result, \textit{italic} is second best.}
	\centering
	\begin{tabular}{ lcCCC@{\extracolsep{10pt}}CCC }
    	\toprule
         \multirow{2}{6em}{Model} & \multirow{2}{*}{Experiment} & \multicolumn{3}{c}{DTU} & \multicolumn{3}{c}{Forward facing}\\[2pt]
         \cline{3-5}
         \cline{6-8}
    	\rule{0pt}{3ex} & & PSNR$\uparrow$ & SSIM$\uparrow$ & LPIPS$\downarrow$ & PSNR$\uparrow$ & SSIM$\uparrow$ & LPIPS$\downarrow$\\
    	\midrule
    	RegNeRF*~\cite{niemeyer2022regnerf} &  Optimised  &  18.89  & 0.745 & 0.190 & 19.08  & 0.587 & 0.336 \\ 
    	\midrule
    	IBRNet~\cite{wang2021ibrnet} & \multirow{3}{4em}{Unseen} & 12.71 & 0.4772 & 0.5678 & 16.40 & 0.5230 & 0.4986\\%
    	MVSNeRF~\cite{chen2021mvsnerf} & & \textit{18.92} & \textit{0.6831} & \textit{0.2580}  & \textbf{16.98} & \textbf{0.5839} & \textit{0.3853}\\
    	Ours &    &  \textbf{19.03}  & \textbf{0.6929} & \textbf{0.2066} & \textit{16.55} & \textit{0.5534} & \textbf{0.3441}\\
    	\bottomrule
	\end{tabular}
	\label{table:quantitative_evaluation}
\end{table}

From table~\ref{table:quantitative_evaluation} we can see that our proposed approach performs similarly to RegNeRF in terms of accuracy. This is despite the fact that RegNeRF is trained in a scene specific regime, while our approach is trained on unrelated scenes, then applied to a completely unknown scene at test time.

\begin{figure}
\begin{tabular}{p{0.165\linewidth}p{0.165\linewidth}p{0.165\linewidth}p{0.165\linewidth}p{0.165\linewidth}}
 Ref. view 1 & Ref. view 2 & GT Target & Pred. Target & Pred. Depth \\
\end{tabular}
\includegraphics[width=\linewidth]{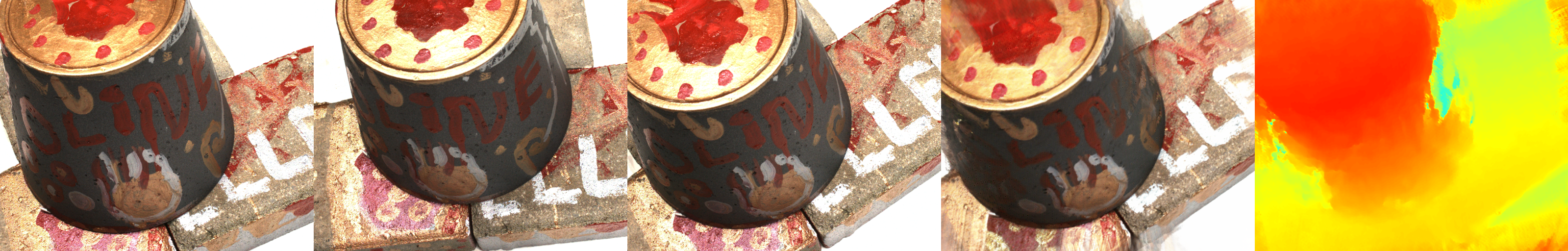}\\
\includegraphics[width=\linewidth]{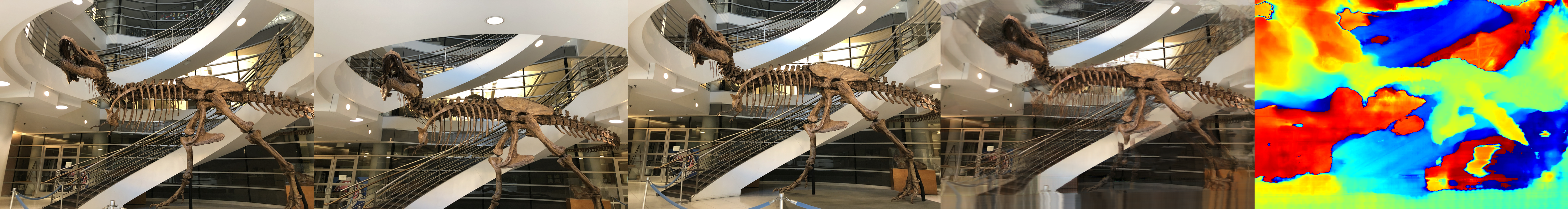}
\vspace{0.1cm}
\caption{Example predictions for the DTU (top) and Forward Facing (bottom) datasets.}
\label{fig:qual}
\vspace{-0.4cm}
\end{figure}

When comparing our technique against the other scene agnostic state-of-the-art approaches (IBRNet and MVSNeRF) under the Sparse View Synthesis evaluation protocol, we note that the simplistic PSNR and SSIM accuracy measures are relatively similar. However, drastic improvements are seen in the LPIPS metric over previous work ranging from a 15\% to 60\% improvement in the perceptual accuracy of the reconstructed scene.
It is interesting to note that our approach performs especially well on the more challenging DTU dataset. Qualitative examples for both datasets are shown in figure~\ref{fig:qual}. For additional examples please see the supplementary material.

\begin{table}[htpb]
    \captionsetup{skip=3pt,width=.9\textwidth}
	\caption{\textbf{Ablation study.} We study the effect of each addition to the model on the Forward facing dataset. \textbf{Bold} is best result, \textit{italic} is second best.}
	\centering
	\begin{tabular}{ ccc|CCC }
    	\toprule
    	Depth Smooth & Distortion & Adversarial & PSNR$\uparrow$ & SSIM$\uparrow$ & LPIPS$\downarrow$\\
    	\midrule
    	$\times$ & $\times$ & $\times$     & 16.11 & 0.5318 & 0.4791\\ %
    	$\checked$ & $\times$ & $\times$     & \textit{16.14} & 0.5361 & 0.4709\\ 
    	$\checked$ & $\checked$ & $\times$     & 16.10 & \textit{0.5419} & \textit{0.4611}\\ 
    	$\checked$ & $\checked$ & $\checked$     & \textbf{16.55} & \textbf{0.5534} & \textbf{0.3441}\\ %
    	\bottomrule
	\end{tabular}
	\label{table:ablation_study}
\end{table}


\subsection{Ablation study}

In table~\ref{table:ablation_study} we undertake an ablation study on the Forward Facing dataset. The depth smoothing loss makes a small but noticable difference across all metrics. It is interesting to note that the distortion loss leads to a marginal decrease in the PSNR and SSIM metrics. However, it provides a more significant improvement in terms of LPIPS. This is expected, as the distortion loss slightly limits the flexibility of the volumetric rendering by preventing ``smearing'' the scene across depths. However, this in turn removes floating blob artifacts and blurred scene depth when viewed from distant viewpoints. These artifacts have a significant impact in the overall perceptual quality of the rendered image, and are vital for possible human-centric applications.

The final adversarial loss leads to significant improvements across all metrics, with the largest gains once again being with the LPIPS score. This demonstrates that the integration of adversarial learning is vital for producing plausible renders for Sparse View Synthesis.

\section{Conclusions}
In this paper we have proposed the Sparse View Synthesis problem. This is a view synthesis problem where the number of reference views is limited, and the baseline between target and reference view is significant. This is a common scenario in live event capture, virtual reality and similar domains.

This imposes a number of challenges which are not present in the standard novel view synthesis problem setup. Most notably the fact that large portions of the target view may be occluded or otherwise not visible within the reference views. In this case there is no mechanism for a standard radiance field model to appropriately fill the gap.

Therefore we proposed an algorithm which unified generative adversarial learning techniques with traditional radiance field modelling. This encouraged the system to inpaint unobserved regions with plausible scene completions. This led to perceptual quality improvements of up to 60\% compared to existing radiance field models.

Nonetheless, there is still some way to go to achieve full extreme Sparse View Synthesis. Although GANs produce good results at generating new content, they suffer from the classic training instability, which makes the model harder to train. In addition, the difficulty of the problem means the complexity of the solution increases. As we handle extreme baseline movements, this creates an ill-posed problem where sometimes the network doesn’t differentiate between empty or occluded space. Thus, in areas viewed by only one of the source views, the reconstruction can lack fidelity. In future work it may be possible to resolve this by re-weighting the generative losses in different regions based on visibility. Regardless, the proposed approach is a major step towards achieving more extreme and sparse renderings.

In future work, it would be interesting to explore the integration of alternative generative modelling techniques with radiance field models. In particular, if the radiance field is able to recognise areas in which it is uncertain, diffusion networks could inpaint these regions directly.

\paragraph{Acknowledgements.}This work was partially supported by the British Broadcasting Corporation (BBC) and the Engineering and Physical Sciences Research Council's (EPSRC) industrial CASE project ``Generating virtual camera views with generative networks'' (voucher number 19000033).

\bibliography{vmgbib}

\newpage

\newcommand\imwidth{0.19\textwidth}
\begin{figure}[!htb]
    \centering
    \setlength{\tabcolsep}{1pt}
    \begin{tabular}{cccccc}
     \rotatebox{90}{Ref. view 1} & \includegraphics[width=\imwidth]{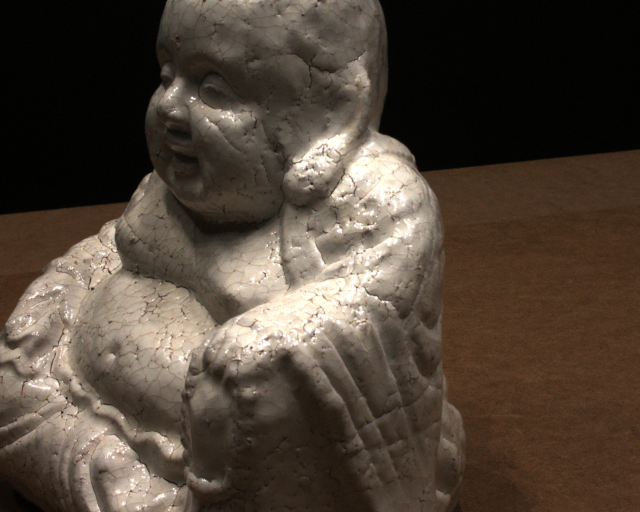} &
     \includegraphics[width=\imwidth]{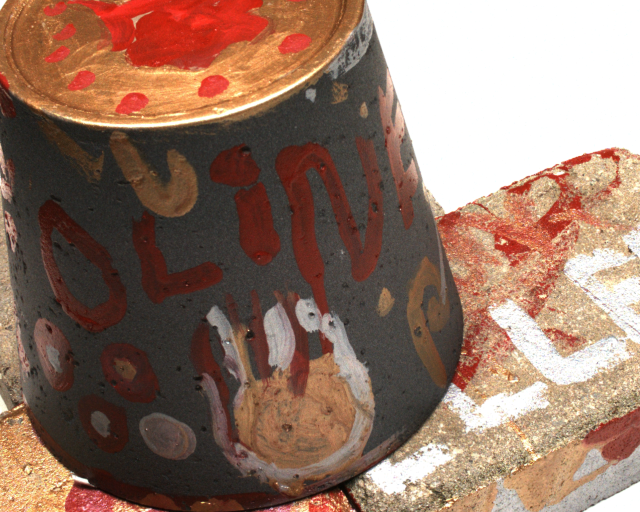}&
     \includegraphics[width=\imwidth]{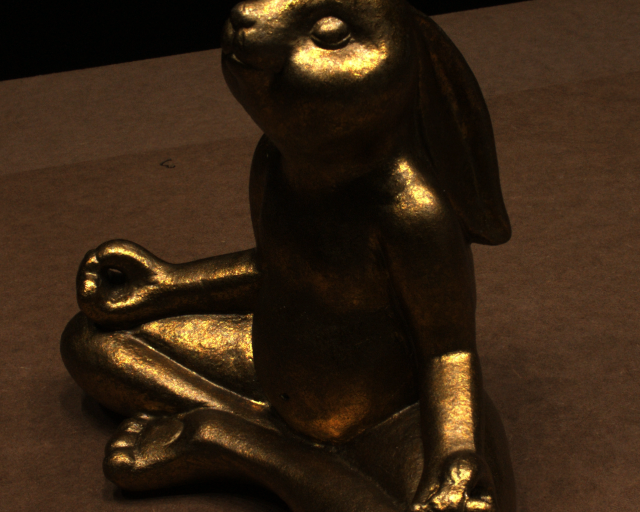} &
     \includegraphics[width=\imwidth]{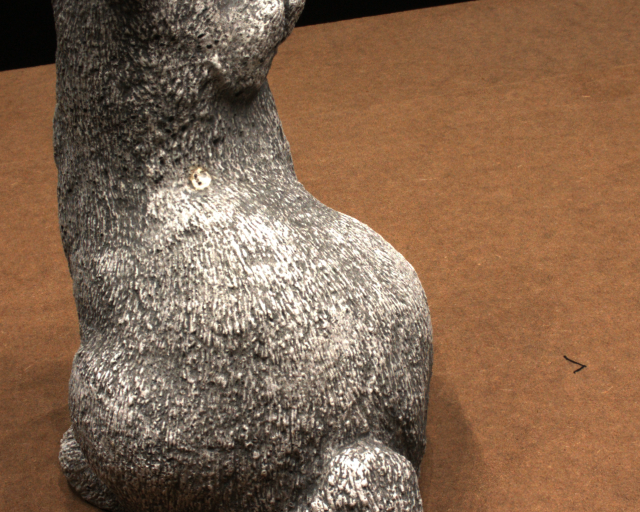}&
     \includegraphics[width=\imwidth]{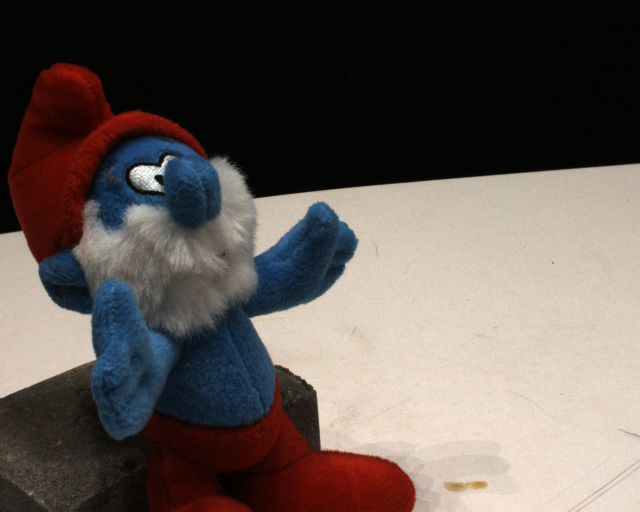}\\
     \rotatebox{90}{Ref. view 2} & \includegraphics[width=\imwidth]{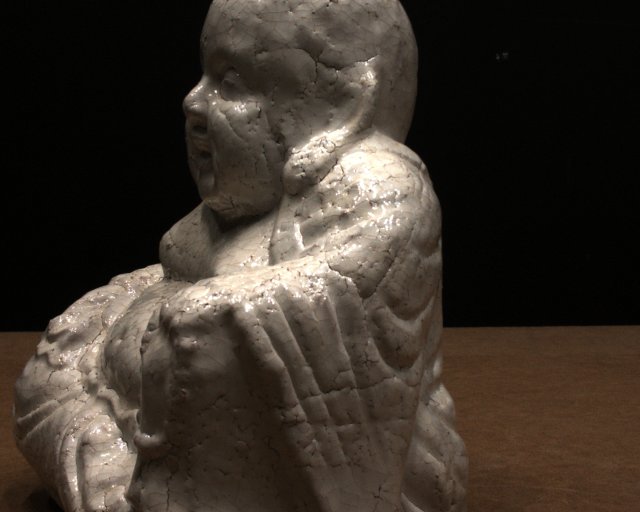} &
     \includegraphics[width=\imwidth]{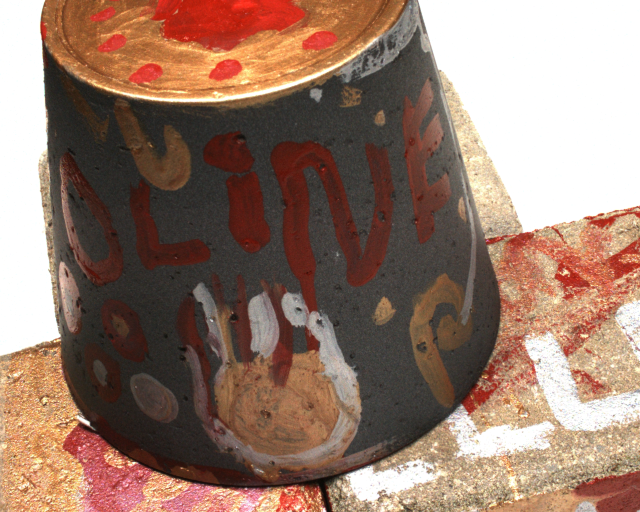}&
     \includegraphics[width=\imwidth]{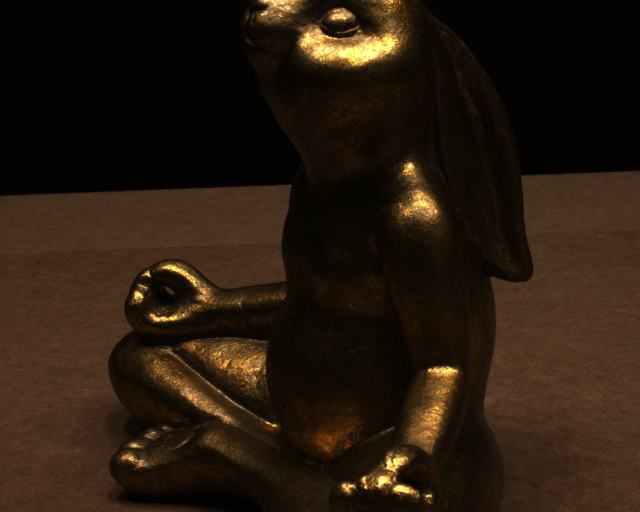} &
     \includegraphics[width=\imwidth]{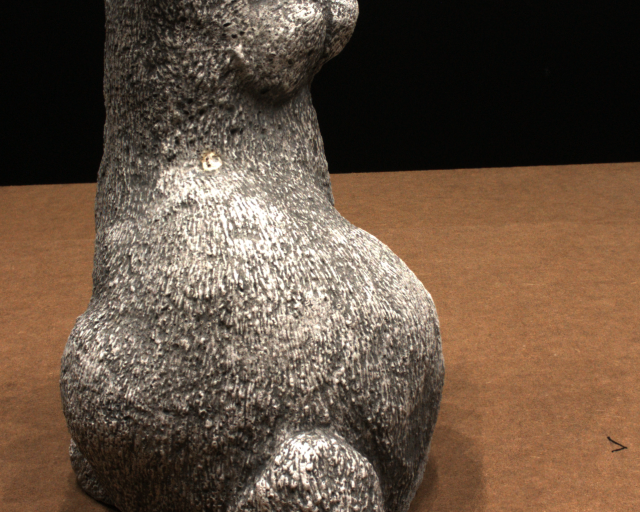}&
     \includegraphics[width=\imwidth]{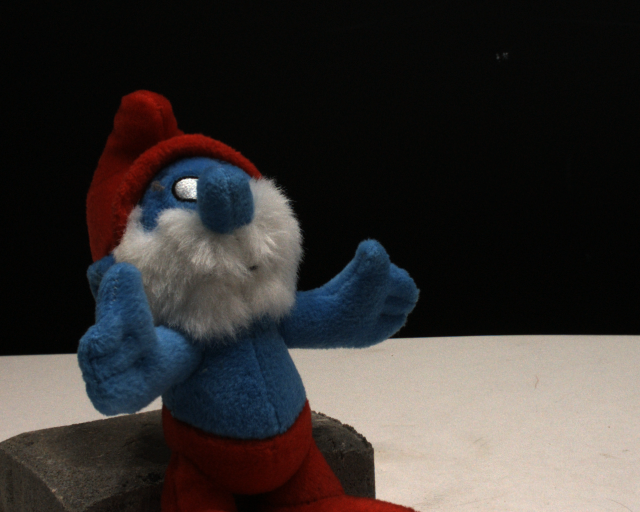}\\
     \rotatebox{90}{MVSNeRF\cite{chen2021}} & \includegraphics[width=\imwidth]{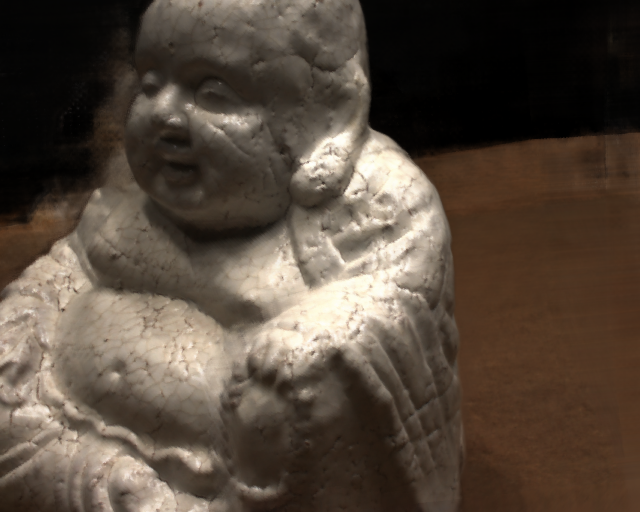} &
     \includegraphics[width=\imwidth]{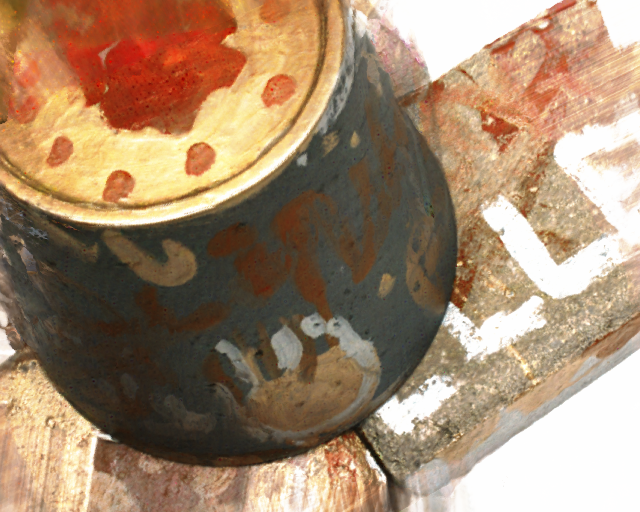}&
     \includegraphics[width=\imwidth]{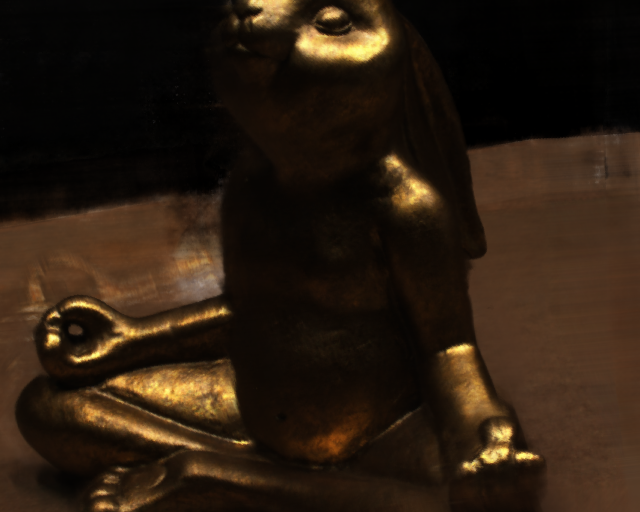} &
     \includegraphics[width=\imwidth]{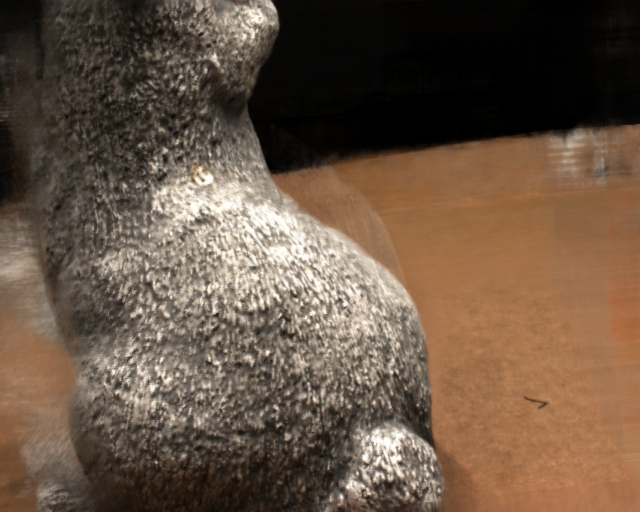}&
     \includegraphics[width=\imwidth]{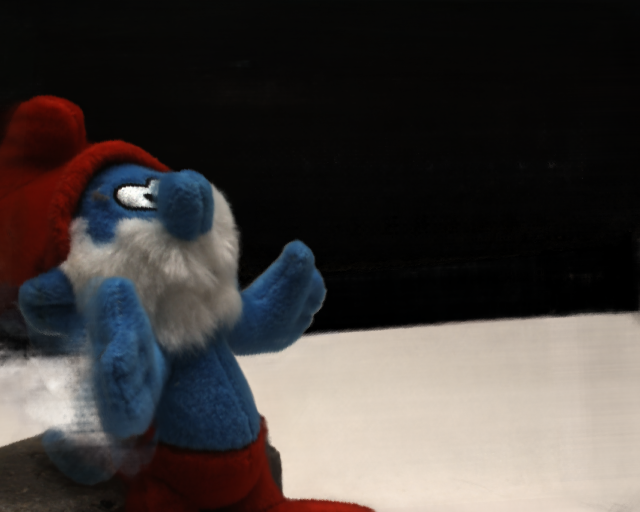} \\
     \rotatebox{90}{Ours} & \includegraphics[width=\imwidth]{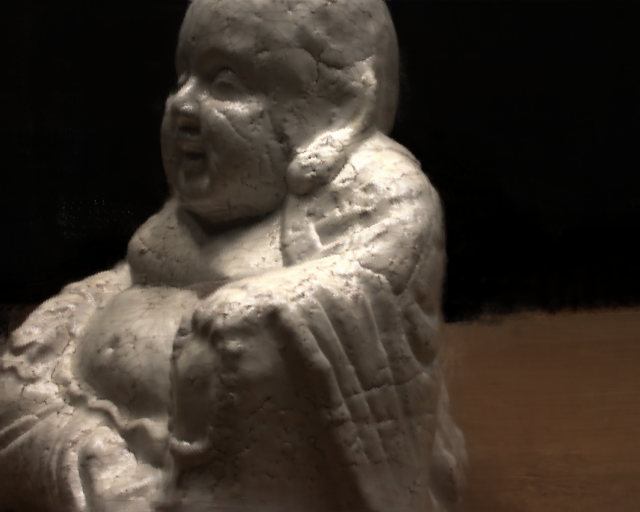} &
     \includegraphics[width=\imwidth]{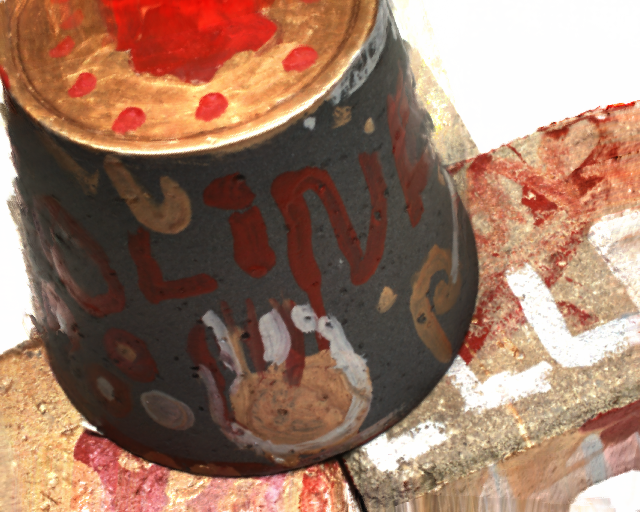}&
     \includegraphics[width=\imwidth]{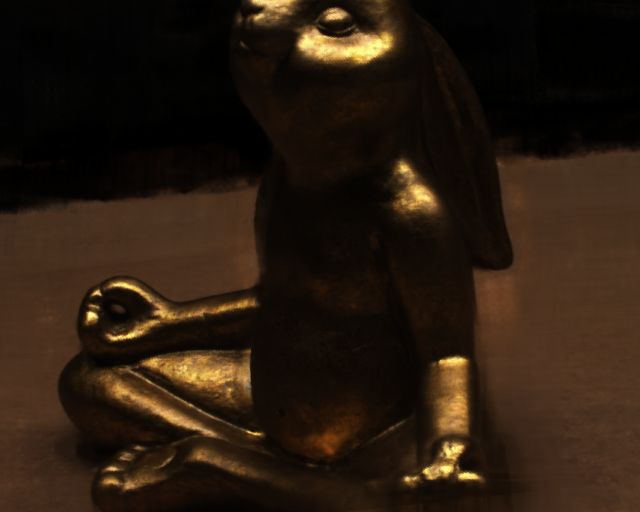} &
     \includegraphics[width=\imwidth]{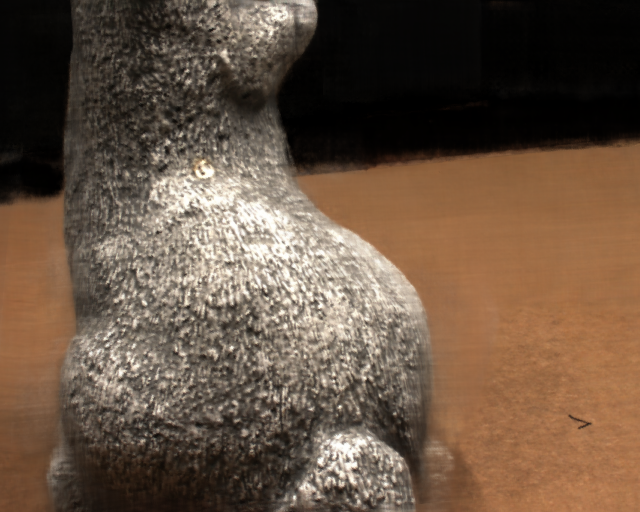}&
     \includegraphics[width=\imwidth]{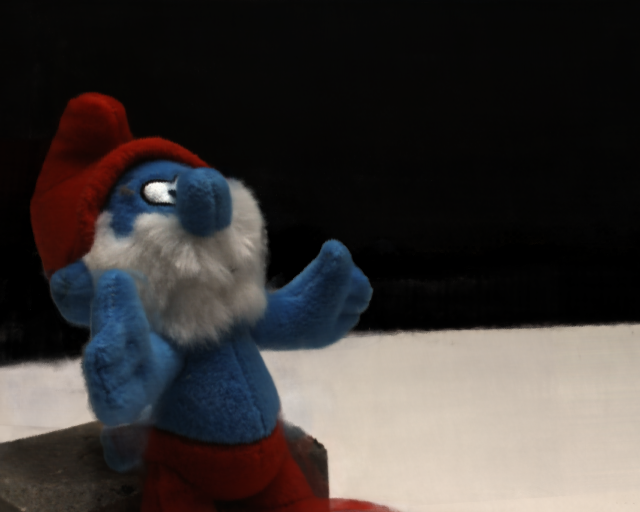}\\
     \rotatebox{90}{Ground Truth} & \includegraphics[width=\imwidth]{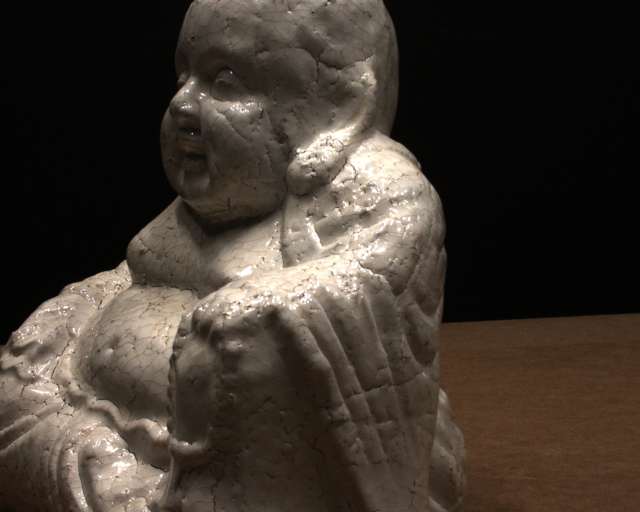} &
     \includegraphics[width=\imwidth]{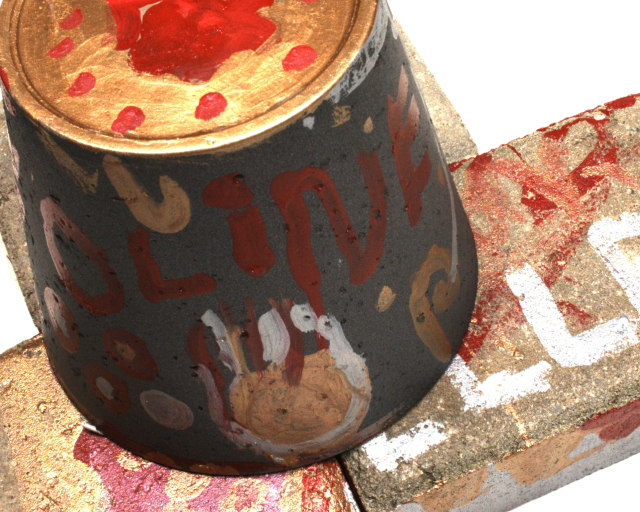}&
     \includegraphics[width=\imwidth]{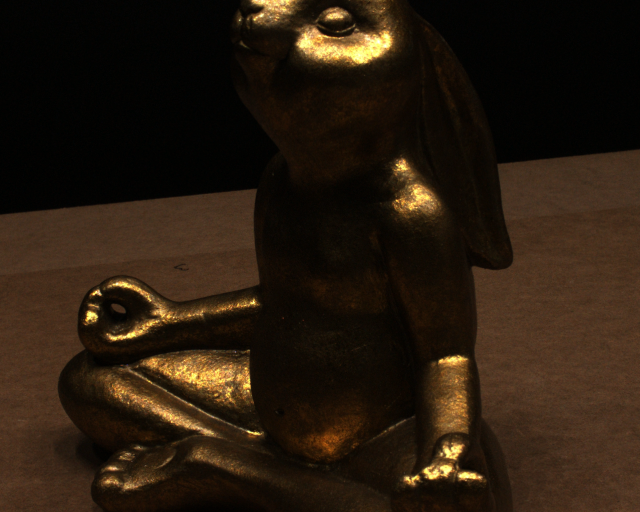} &
     \includegraphics[width=\imwidth]{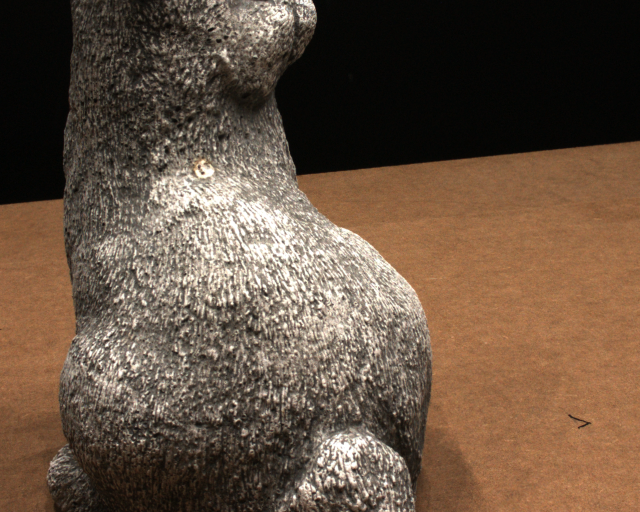}&
     \includegraphics[width=\imwidth]{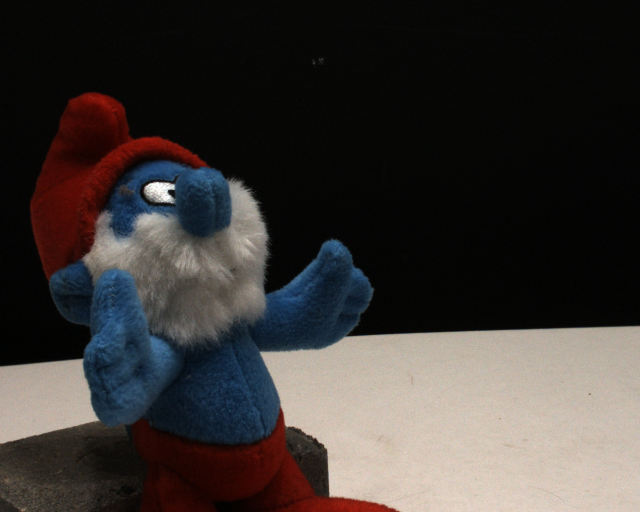} \\
     \rotatebox{90}{Depth (Ours)} & \includegraphics[width=\imwidth]{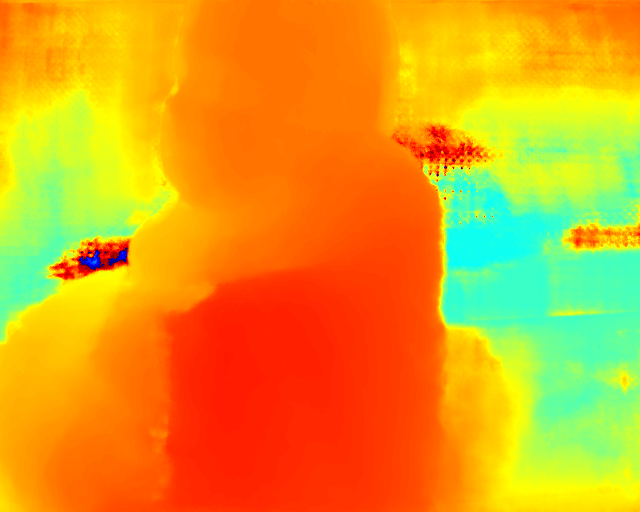} &
     \includegraphics[width=\imwidth]{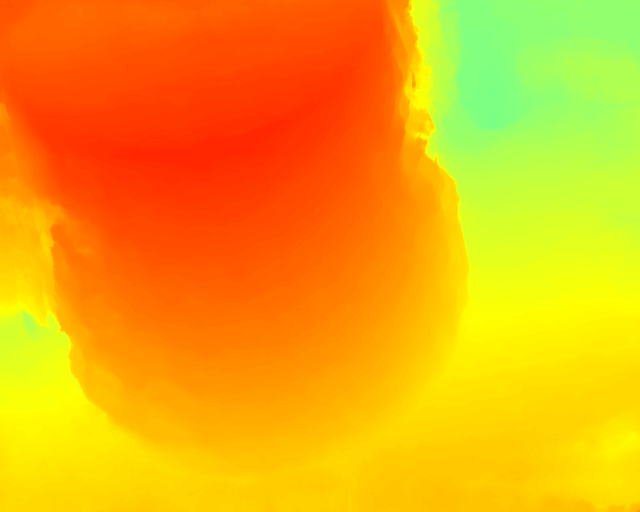}&
     \includegraphics[width=\imwidth]{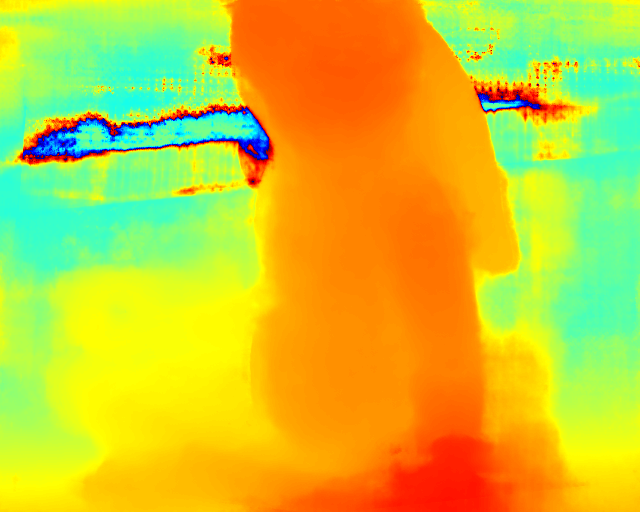} &
     \includegraphics[width=\imwidth]{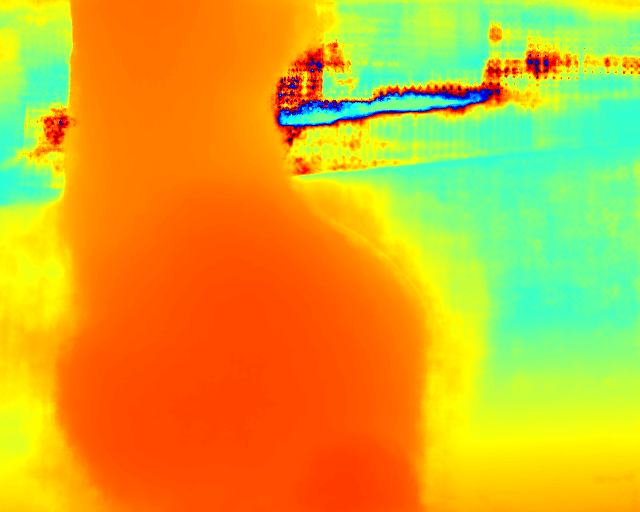}&
     \includegraphics[width=\imwidth]{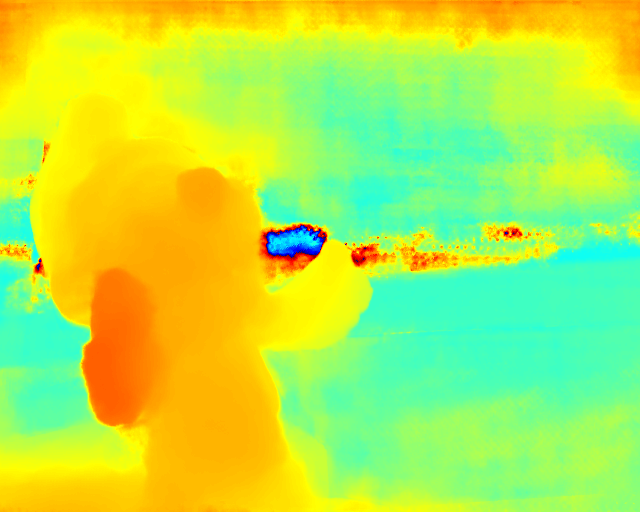} \\
    \end{tabular}
    \vspace{0.5cm}
    \caption{Example predictions from the DTU dataset. Each column shows a different scene predicted under the Sparse View Synthesis evaluation protocol. Note that state of the art \cite{chen2021mvsnerf} predictions exhibit fuzziness or ``smearing'' around object boundaries where occlusions occur. In contrast our approach provides sharper edges as it realistically fills unobserved regions.}
    \label{fig:dtu}
\end{figure}

\newcommand\imwidthb{0.24\textwidth}
\begin{figure}[!htb]
    \centering
    \setlength{\tabcolsep}{1pt}
    \begin{tabular}{ccccc}
     \rotatebox{90}{Ref. view 1} &
     \includegraphics[width=\imwidthb]{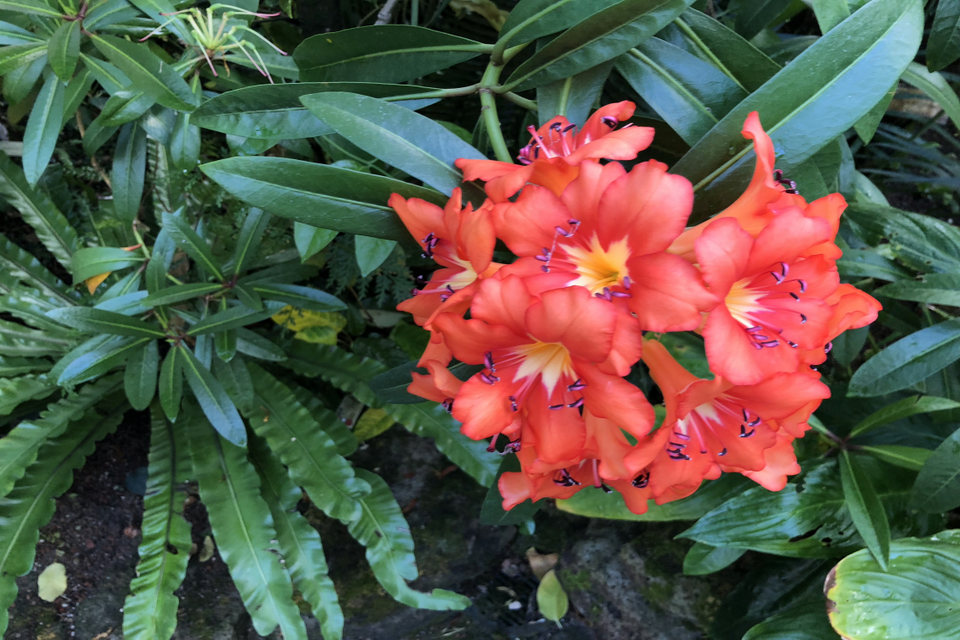}&
     \includegraphics[width=\imwidthb]{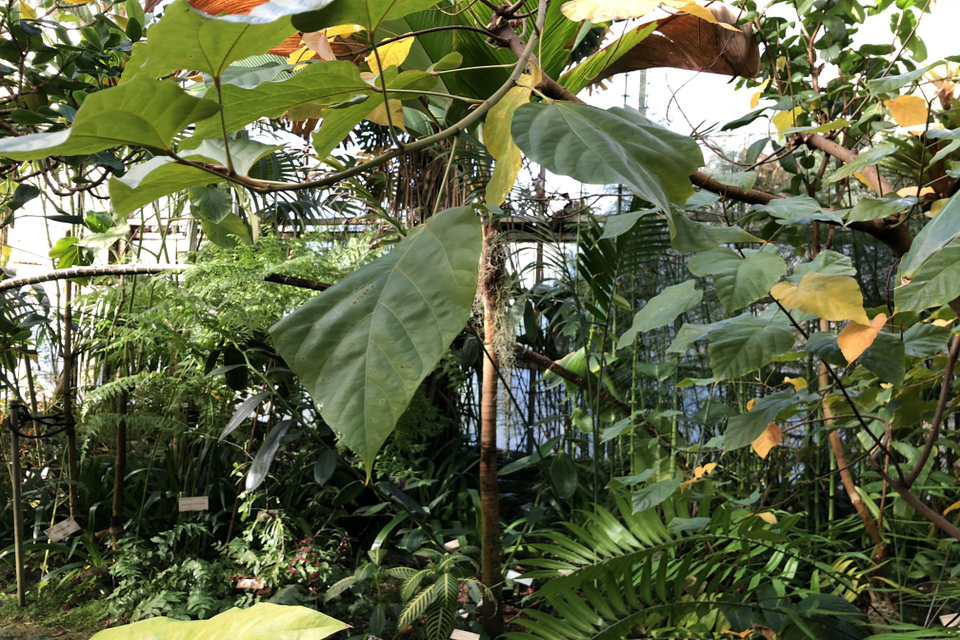} &
     \includegraphics[width=\imwidthb]{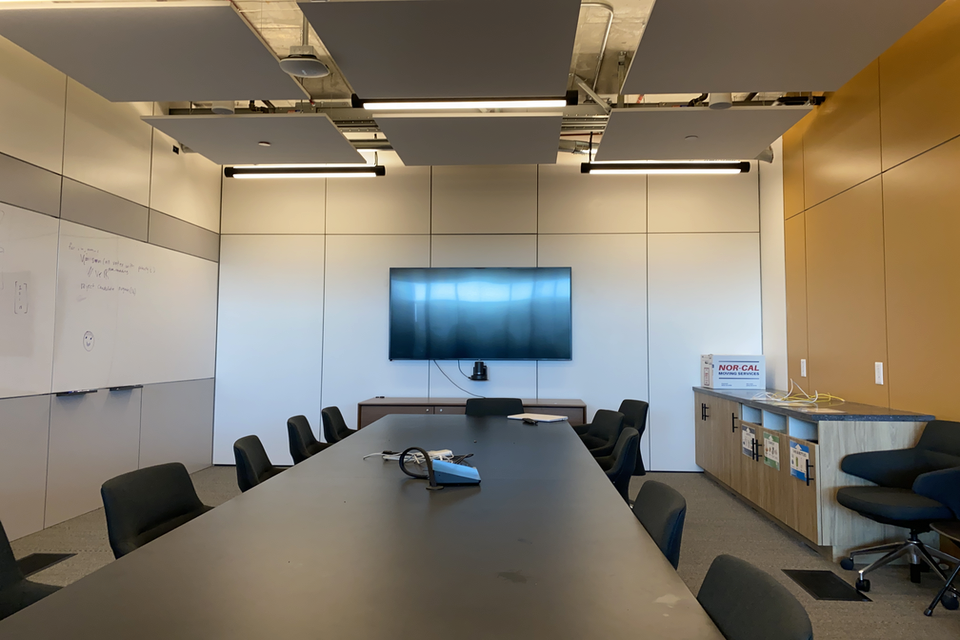}&
     \includegraphics[width=\imwidthb]{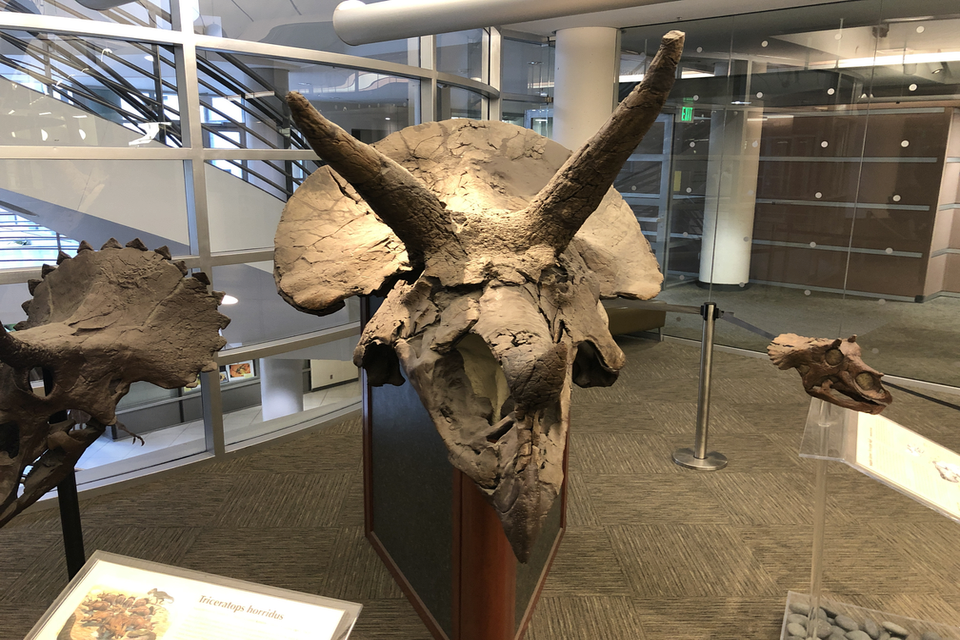}\\
     \rotatebox{90}{Ref. view 2} &
     \includegraphics[width=\imwidthb]{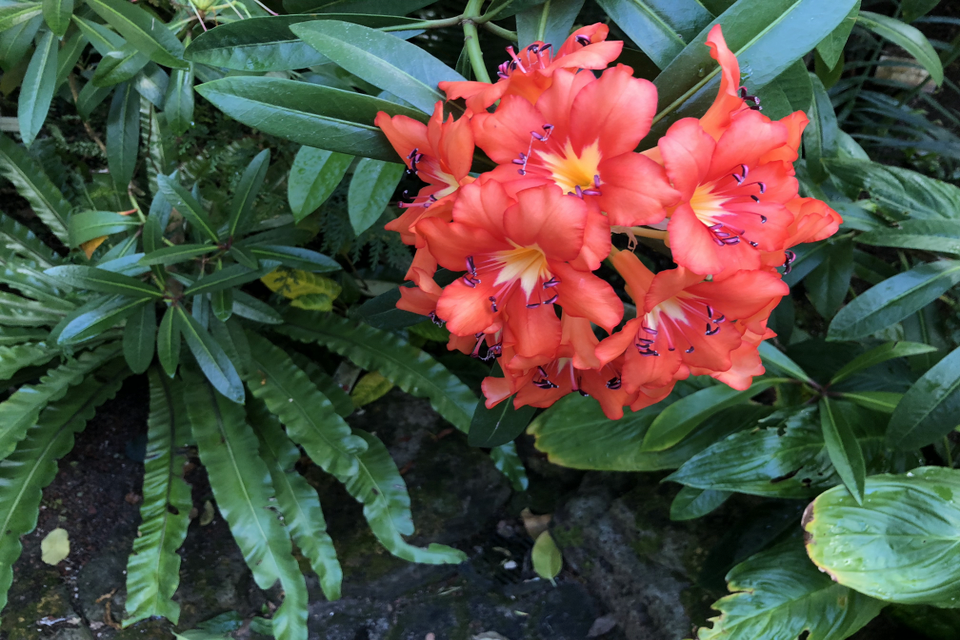}&
     \includegraphics[width=\imwidthb]{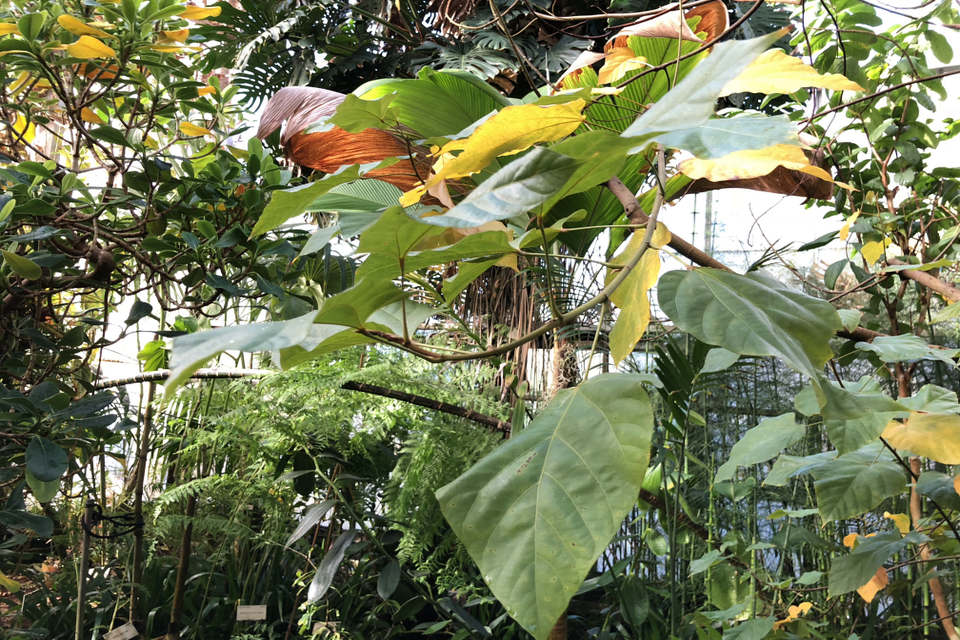} &
     \includegraphics[width=\imwidthb]{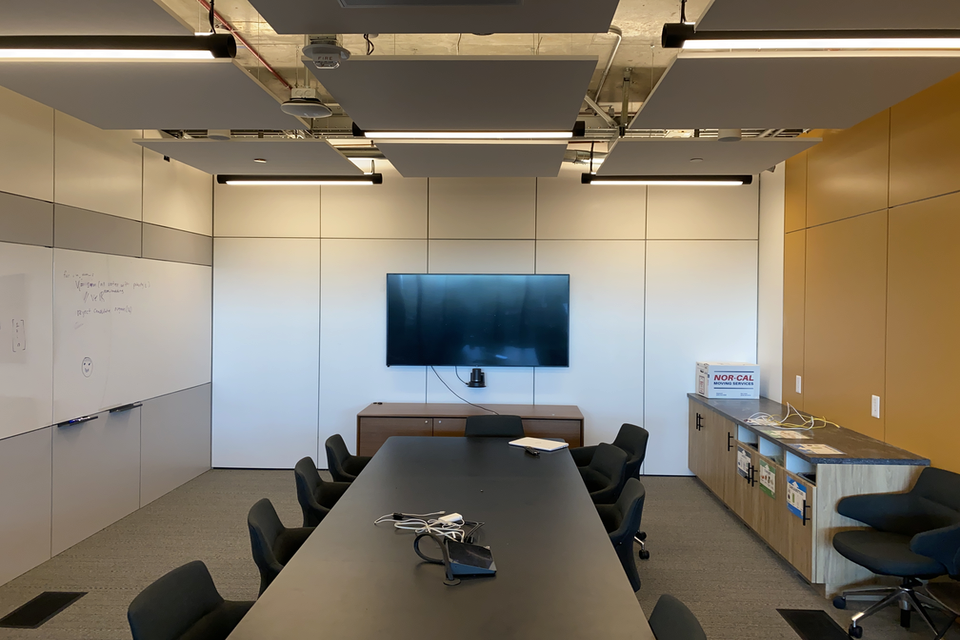}&
     \includegraphics[width=\imwidthb]{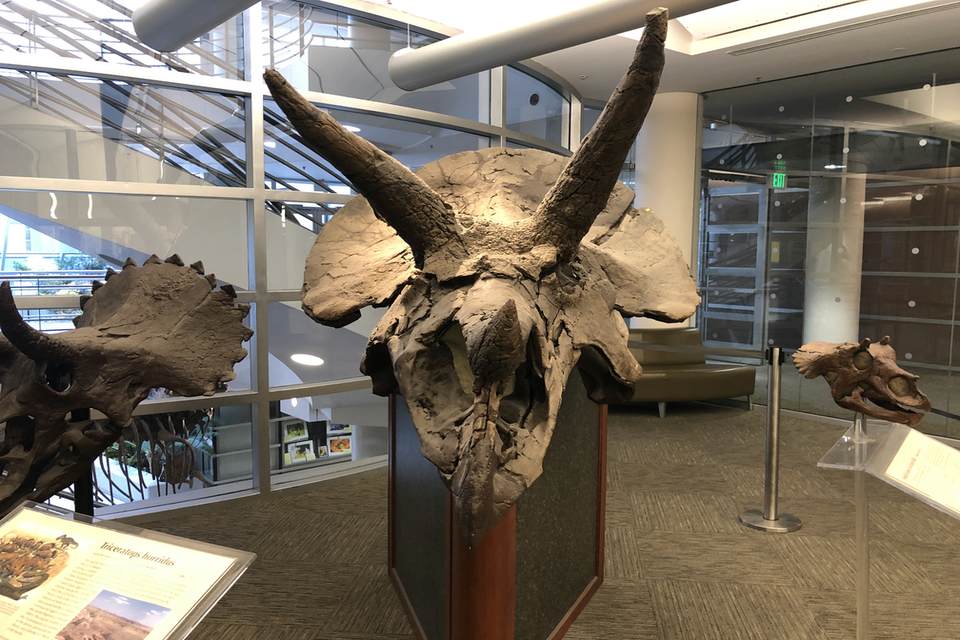}\\
     \rotatebox{90}{MVSNeRF\cite{chen2021mvsnerf}} &
     \includegraphics[width=\imwidthb]{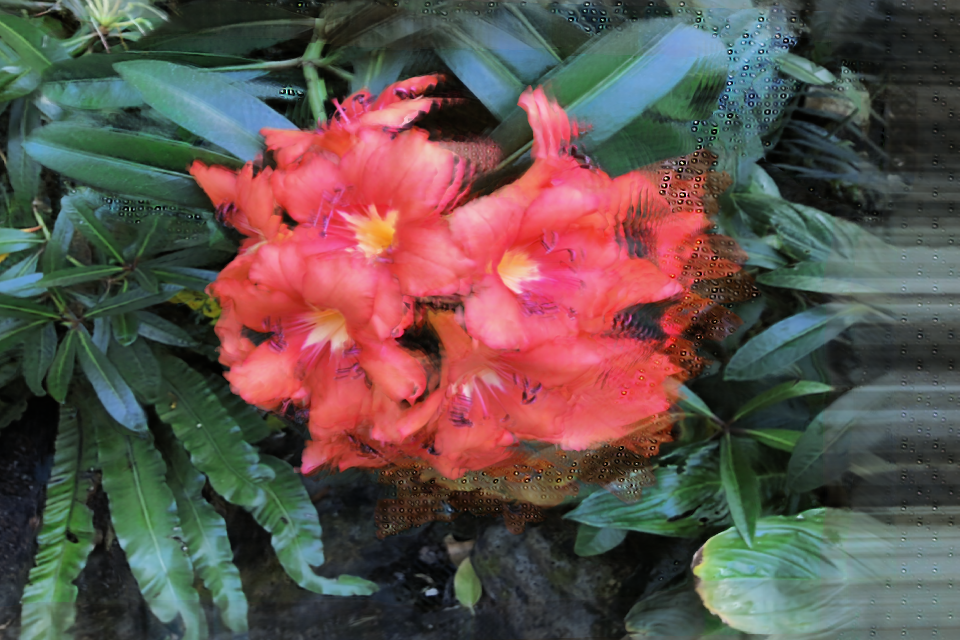}&
     \includegraphics[width=\imwidthb]{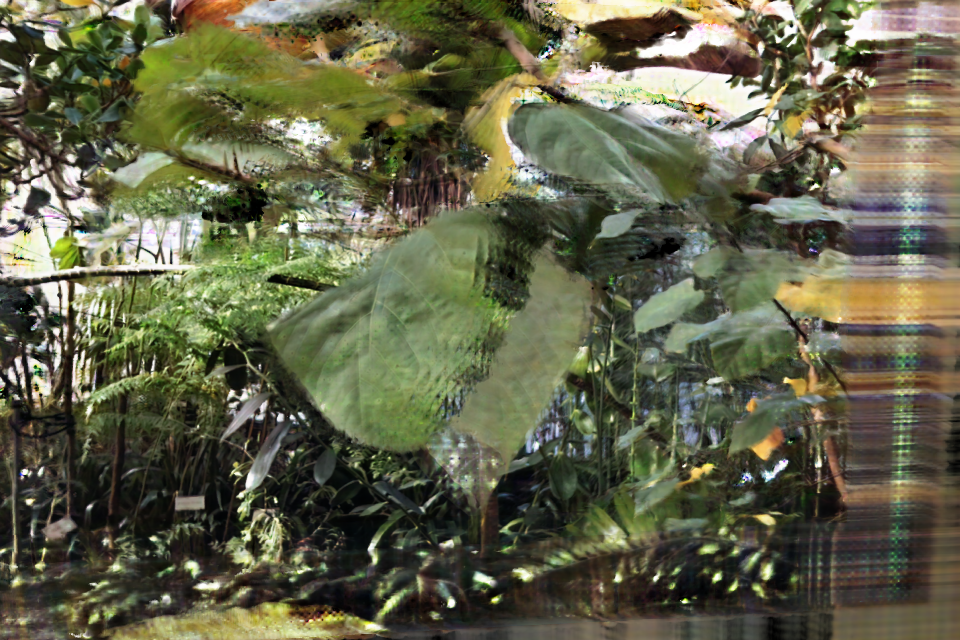} &
     \includegraphics[width=\imwidthb]{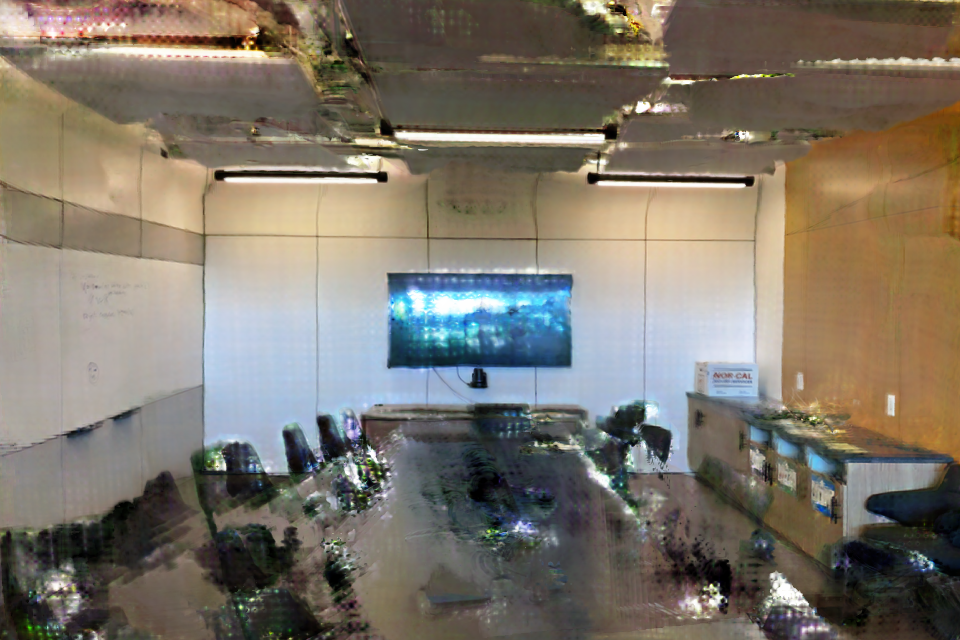}&
     \includegraphics[width=\imwidthb]{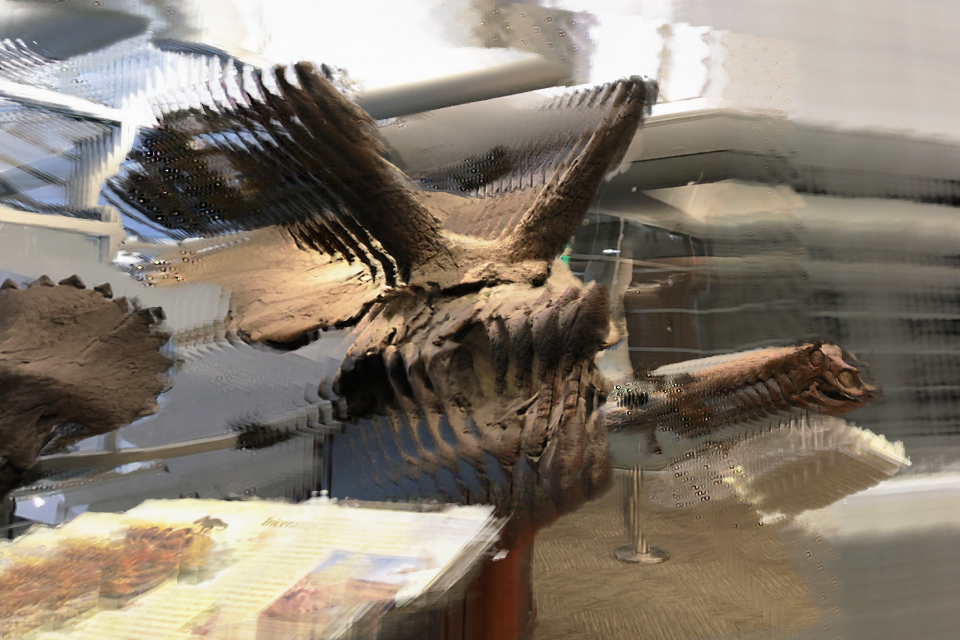} \\
     \rotatebox{90}{Ours} &
     \includegraphics[width=\imwidthb]{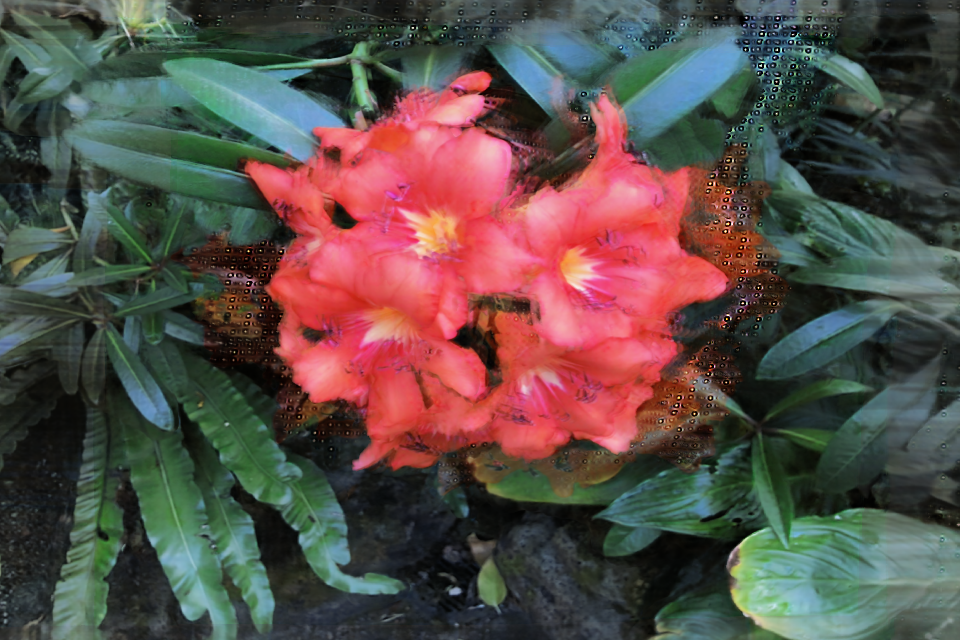}&
     \includegraphics[width=\imwidthb]{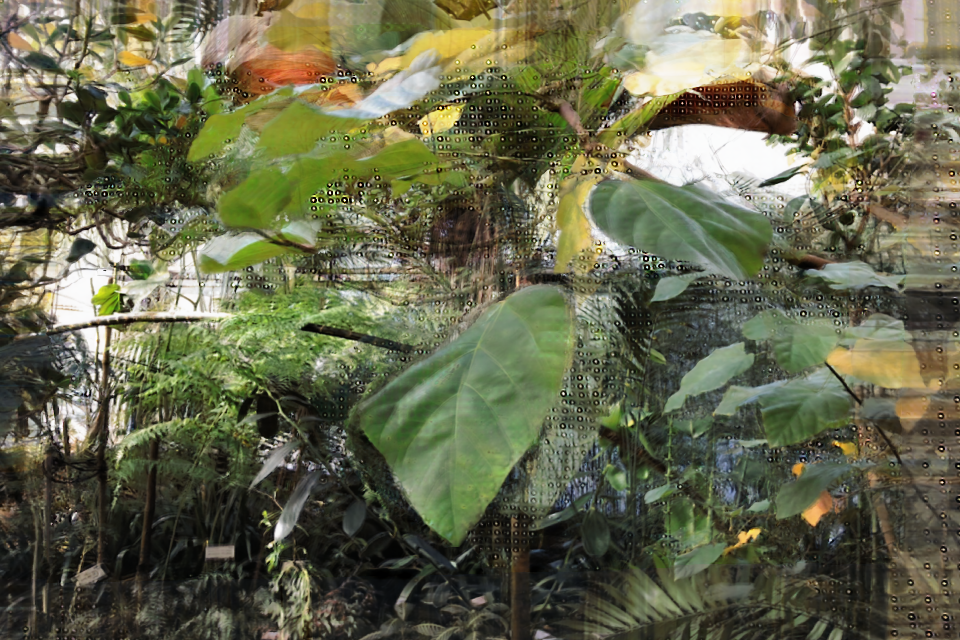} &
     \includegraphics[width=\imwidthb]{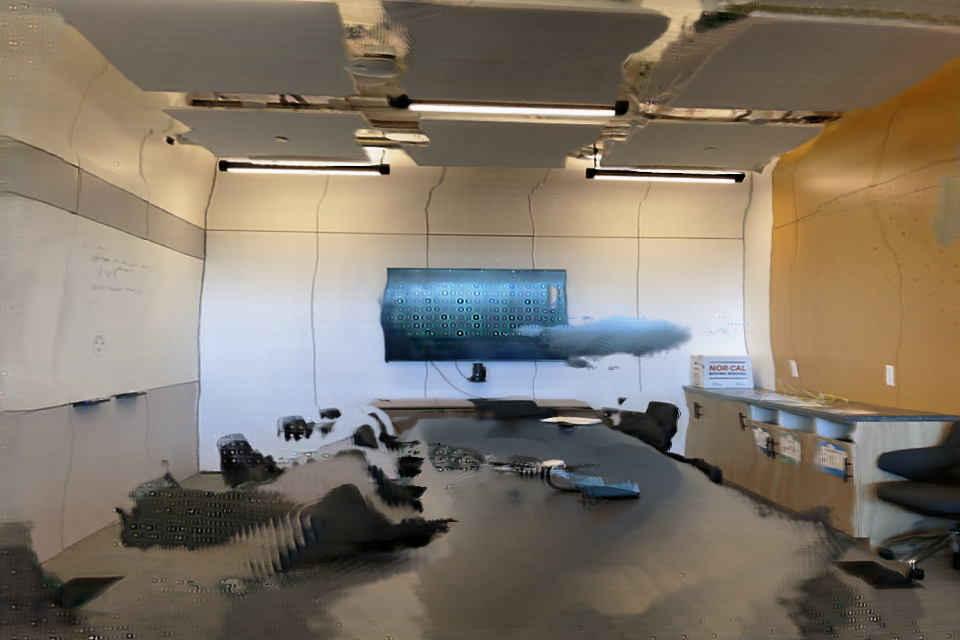}&
     \includegraphics[width=\imwidthb]{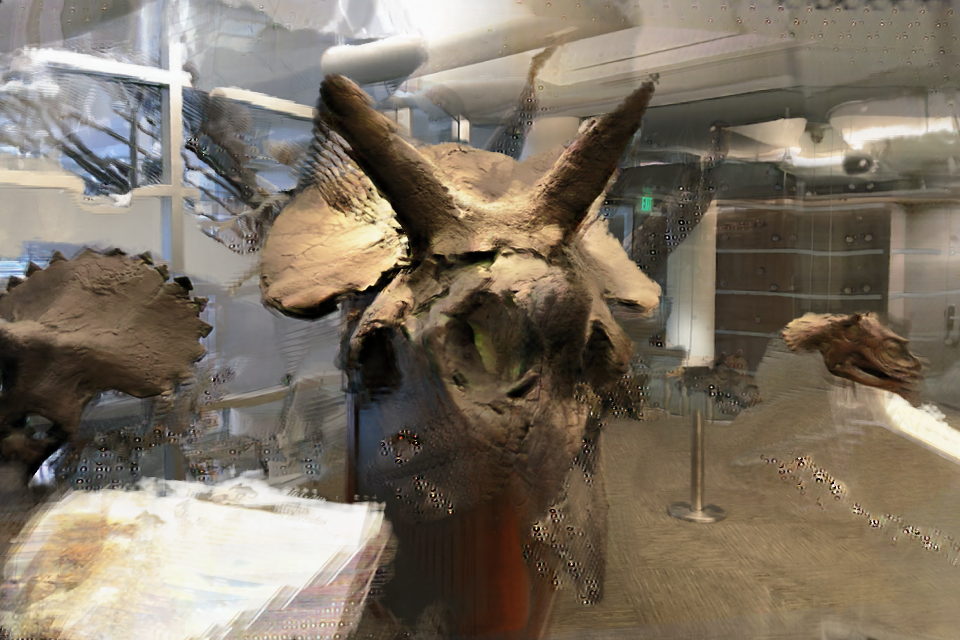}\\
     \rotatebox{90}{Ground Truth} &
     \includegraphics[width=\imwidthb]{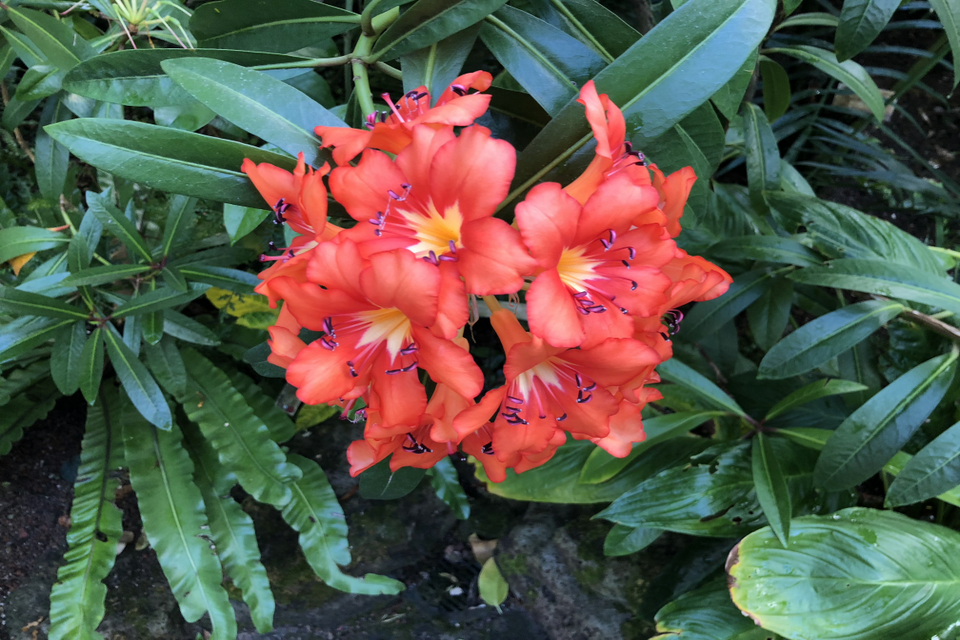}&
     \includegraphics[width=\imwidthb]{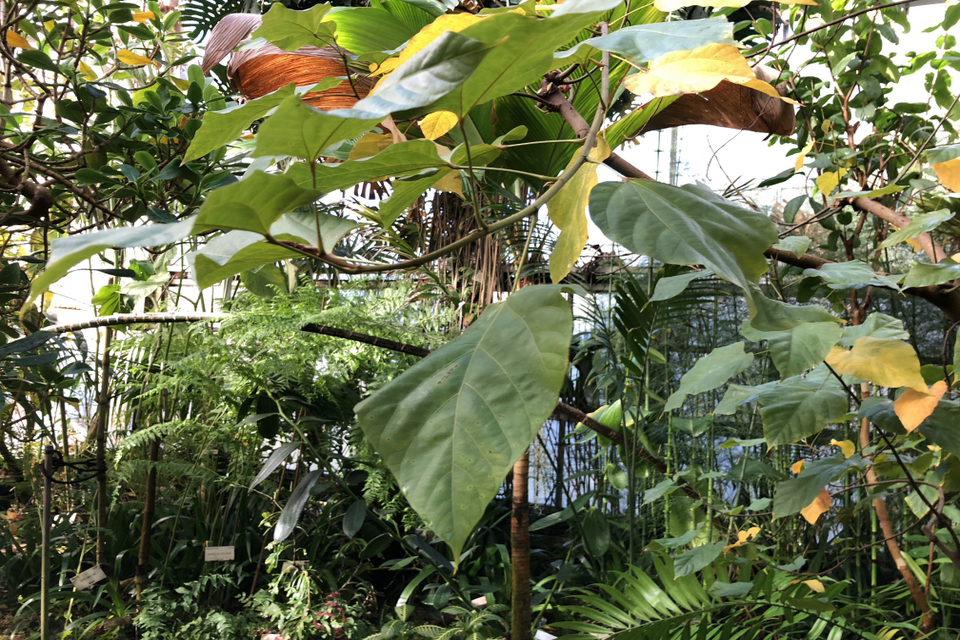} &
     \includegraphics[width=\imwidthb]{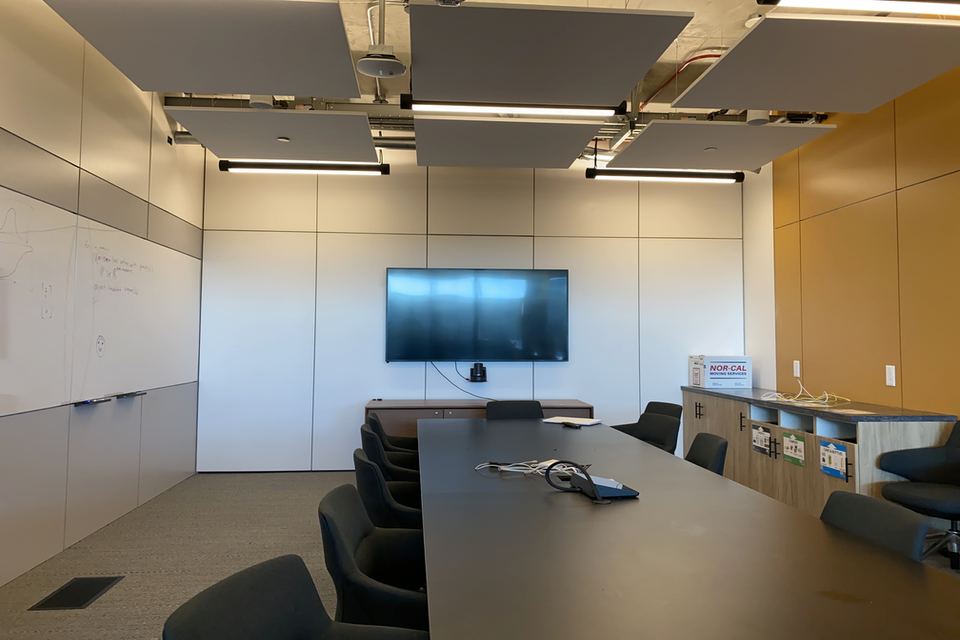}&
     \includegraphics[width=\imwidthb]{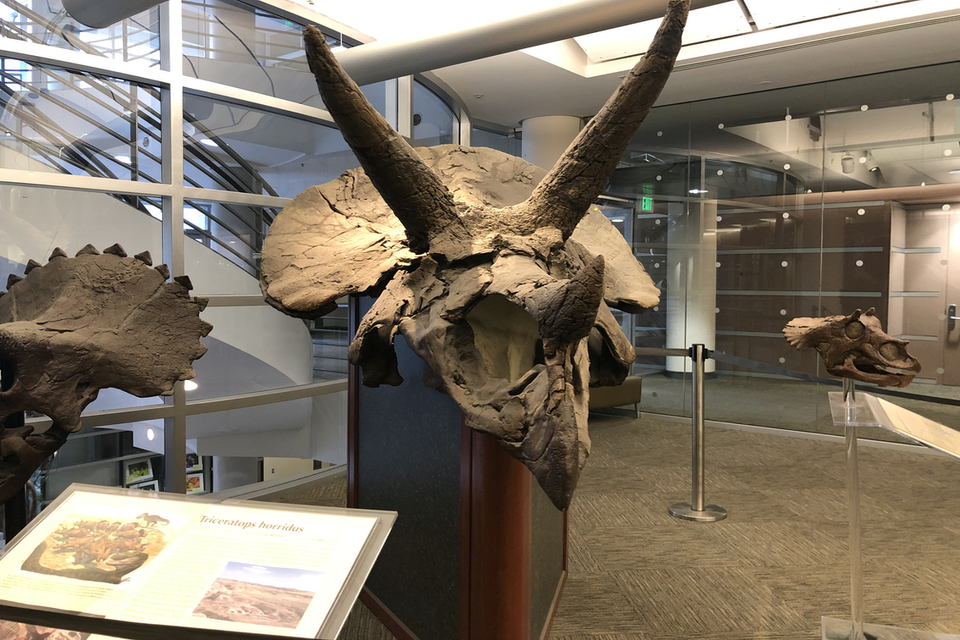} \\
     \rotatebox{90}{Depth (us)} &
     \includegraphics[width=\imwidthb]{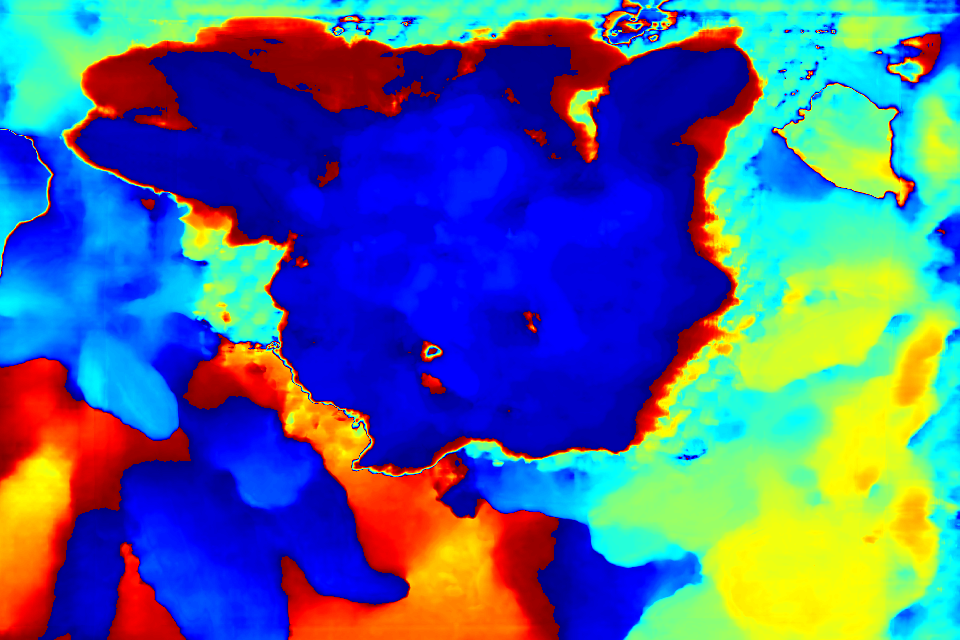}&
     \includegraphics[width=\imwidthb]{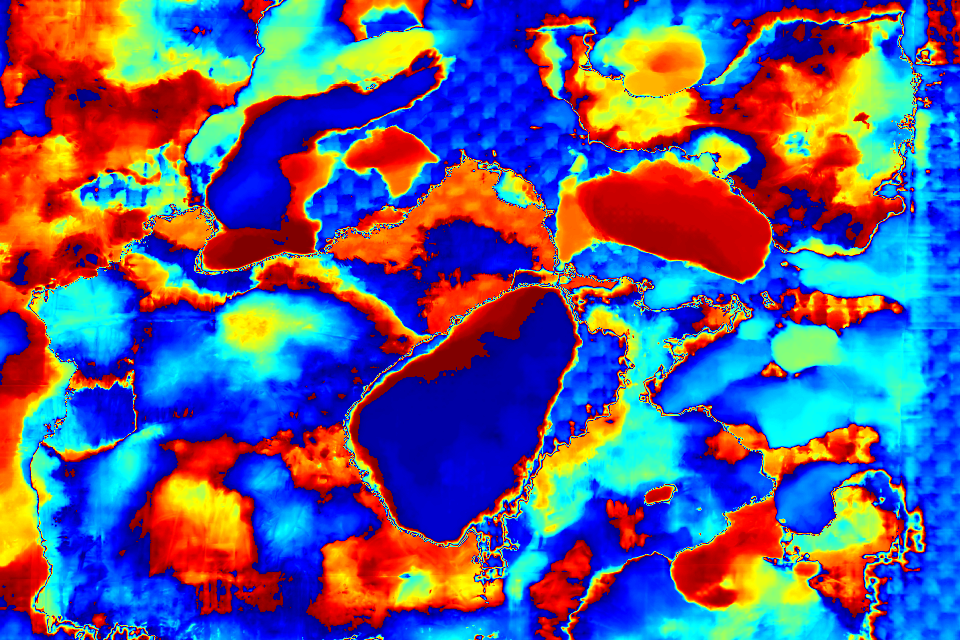} &
     \includegraphics[width=\imwidthb]{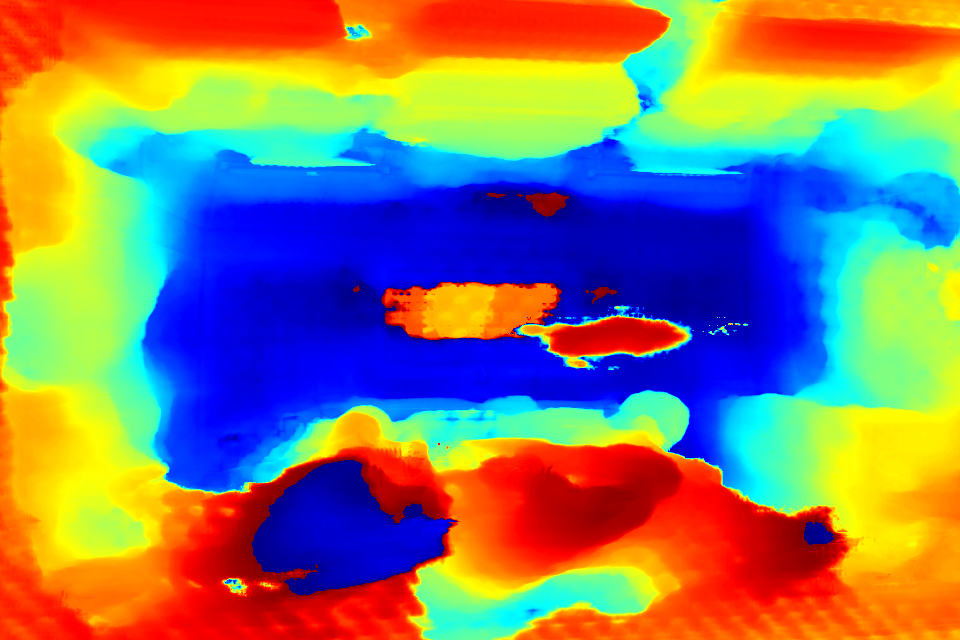}&
     \includegraphics[width=\imwidthb]{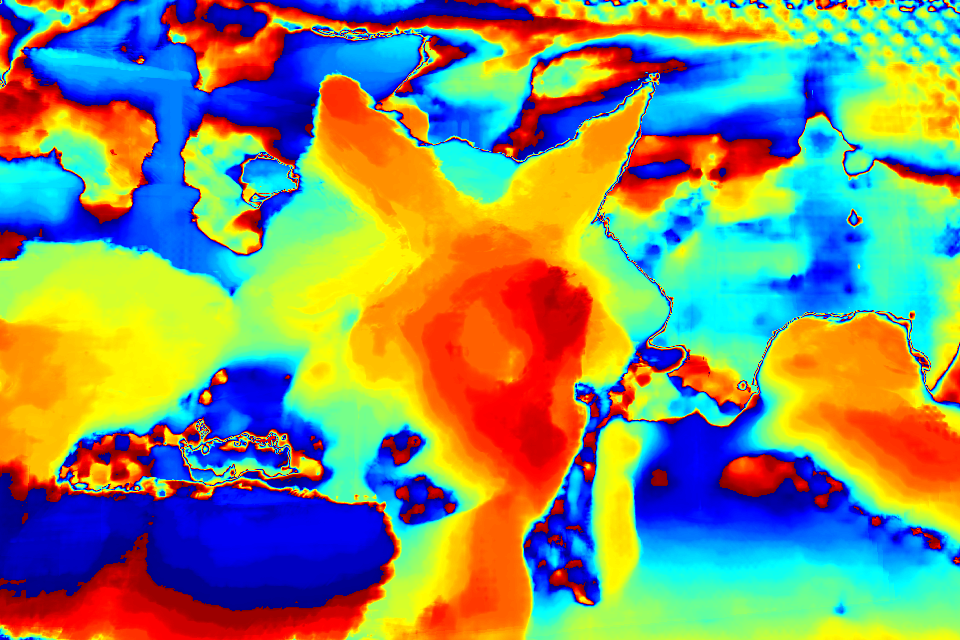} \\
    \end{tabular}
    \vspace{0.5cm}
    \caption{Example predictions from the LLFF dataset. Again we can observe ``smearing'' artifacts in the state-of-the-art predictions, especially along the image borders. Additionally, we can observe ``halo'' or ``duplicated boundary'' artifacts in some of the state-of-the-art scenes such as the rightmost two columns. These perceptually jarring artifacts are easily detected by our adversarial discriminator, and so they do not occur in the predictions for our approach.}
    \label{fig:llff}
\end{figure}

\end{document}